\documentclass[letterpaper]{article} 
\usepackage{aaai2026}  
\usepackage{times}  
\usepackage{helvet}  
\usepackage{courier}  
\usepackage[hyphens]{url}  
\usepackage{graphicx} 
\usepackage[numbers]{natbib}  
\usepackage{caption} 
\usepackage[utf8]{inputenc}
\usepackage{booktabs}     
\usepackage{soul} 
\usepackage{siunitx}      
\usepackage{caption}
\usepackage{threeparttable} 

\usepackage{etoc}
\usepackage{appendix}
\usepackage[dvipsnames]{xcolor}
\usepackage{xcolor}

\usepackage{hyperref}
\hypersetup{
  colorlinks=true,
  linkcolor=MidnightBlue,
  filecolor=magenta,      
  urlcolor=blue,
  pdftitle={Overleaf Example},
  pdfpagemode=FullScreen,
  }

\usepackage{algorithm}
\usepackage{algorithmic}

\usepackage{newfloat}
\usepackage{listings}
\DeclareCaptionStyle{ruled}{labelfont=normalfont,labelsep=colon,strut=off} 
\floatstyle{ruled}
\newfloat{listing}{tb}{lst}{}
\floatname{listing}{Listing}
\usepackage{svg}
\usepackage{booktabs}
\usepackage{multirow}
\usepackage{amsmath}
\usepackage{amssymb}
\usepackage{dsfont}
\usepackage{subcaption}
\usepackage{booktabs}
\usepackage{adjustbox}
\usepackage{threeparttable}
\usepackage{array}
\usepackage{hyperref}
\usepackage{longtable}
\usepackage{booktabs}
\usepackage{hyperref}
\usepackage{array}
\usepackage{longtable}
\usepackage{ragged2e}
\usepackage{hyperref}
\usepackage{array}
\usepackage[table]{xcolor}
\usepackage{xcolor}
\title{\textbf{FOS}: A Large-Scale Temporal Graph Benchmark for Scientific Interdisciplinary Link Prediction}

\author {
    Kiyan Rezaee\textsuperscript{\rm 1}\thanks{Corresponding author: \texttt{rezaeeki@msu.edu}},
    Morteza Ziabakhsh\textsuperscript{\rm 2},
    Niloofar Nikfarjam\textsuperscript{\rm 3},
    Mohammad M. Ghassemi\textsuperscript{\rm 1},
    Yazdan Rezaee Jouryabi\textsuperscript{\rm 4},
    Sadegh Eskandari\textsuperscript{\rm 3},
    Reza Lashgari\textsuperscript{\rm 4}
}

\affiliations {
    \textsuperscript{\rm 1}Department of Computer Science and Engineering, Michigan State University\\\
    \textsuperscript{\rm 2}School of Mathematics, Statistics and Computer Science, College of Science, University of Tehran\\
    \textsuperscript{\rm 3}Department of Computer Science, University of Guilan\\
    \textsuperscript{\rm 4}Institute of Medical Science and Technology, Shahid Beheshti University\\
}

\begin{document}

\maketitle

\begin{abstract}
Interdisciplinary scientific breakthroughs mostly emerge unexpectedly, and forecasting the formation of novel research fields remains a major challenge. We introduce \textbf{FOS} (\underline{F}uture  \underline{O}f  \underline{S}cience), a comprehensive time-aware graph-based benchmark that reconstructs annual co-occurrence graphs of 65,027 research sub-fields (spanning 19 general domains) over the period 1827–2024. In these graphs, edges denote the co-occurrence of two fields in a single publication and are timestamped with the corresponding publication year. Nodes are enriched with semantic embeddings, and edges are characterized by temporal and topological descriptors. We formulate the prediction of new field-pair linkages as a temporal link-prediction task, emphasizing the “first-time” connections that signify pioneering interdisciplinary directions. Through extensive experiments, we evaluate a suite of state-of-the-art temporal graph architectures under multiple negative-sampling regimes and show that (i) embedding long-form textual descriptions of fields significantly boosts prediction accuracy, and (ii) distinct model classes excel under different evaluation settings. Case analyses show that top-ranked link predictions on FOS align with field pairings that emerge in subsequent years of academic publications. We publicly release FOS, along with its temporal data splits and evaluation code, to establish a reproducible benchmark for advancing research in predicting scientific frontiers.
\end{abstract}

\begin{links}
    \link{Code}{https://github.com/kiyan-rezaee/future-of-science}
    \link{Data}{https://huggingface.co/datasets/Morteza24/future-of-science}
\end{links}

\section{Introduction}

Scientific innovation is progressing at an increasingly rapid pace, both in terms of speed and scale~\cite{bornmann2015growth, bornmann2021growth}. The number of publications continues to grow swiftly~\cite{krenn2023forecasting}, disciplinary boundaries are becoming more fluid~\cite{mcbean2017blurring}, and collaborations now span a wider range of fields than in previous decades~\cite{newman2024promoting}. These transformations pose a critical question for researchers, funders, and policymakers: which emerging directions will shape the future of science? While traditional scientometric and bibliometric methods remain crucial for reconstructing past trends~\cite{xia2023review, min2021predicting, glanzel2008seven}, they are limited in their ability to predict future, cross-disciplinary breakthroughs. In short, these methods are more effective at charting where science has been than where it is headed.

Recent advances in machine learning, network science, and large-scale knowledge representation have opened new opportunities for forecasting scientific trajectories. For example, Ofer \textit{et al.}~\cite{ofer2024s} combine publication time series, article-type ratios, semantic topic embeddings, and patent signals to predict topic popularity; Krenn \textit{et al.}~\cite{krenn2023forecasting} model research as a concept network and use link prediction to identify nascent concept pairings; and Gu and Krenn~\cite{gu2025forecasting} use a dynamic knowledge graph including millions of papers and citation analysis to show that the eventual impact of scientific ideas can be predicted even before their formal publication. These studies demonstrate considerable promise, but significant limitations remain.

First, many studies target carefully curated corpora or single domains (e.g., biomedicine, AI, or quantum physics), limiting cross-domain generalization~\cite{rzhetsky2015choosing, krenn2023forecasting, krenn2020predicting}. Second, the majority of empirical analyses operate on static or coarse temporal snapshots, which obscures the fine-grained, long-horizon dynamics by which fields interact and recombine~\cite{ofer2024s, clauset2017data}. Third, existing frameworks~\cite{gu2025forecasting, ofer2024s}  do not explicitly model that \emph{under what conditions} novel interdisciplinary connections arise: sparse historical records, and limited representation of interaction mechanisms leave these generative processes underexplored. Addressing these gaps requires a large-scale, time-aware benchmark that supports fine-grained temporal modeling of field-level relationships.

In this work, we introduce \textbf{FOS} (\underline{F}uture  \underline{O}f  \underline{S}cience), a large-scale, time-aware, graph-based representation of scientific knowledge that is explicitly designed to study the emergence of interdisciplinary connections. In FOS, nodes denote fields of study, edges denote their co-occurrence within publications, and every edge is timestamped with the publication year. FOS covers 65{,}027 distinct fields across 19 top-level scientific domains and is enriched with semantic node features derived from textual metadata, enabling a fine-grained view of when and how field-level relationships arise and evolve.

We formalize forecasting as a temporal link-prediction task: given the history of field co-occurrences up to time $t$, predict which previously unobserved field pairs will form links at times $t' > t$. Our primary interest is in \emph{novel} interdisciplinary connections (pairs that have not co-occurred by time $t$), since such first-time links frequently signal substantive cross-boundary advances~\cite{cheng2025exploring, sebastian2021boundary}. Concretely, we ask the following research questions:

\begin{enumerate}
    \item To what extent can temporal, relational, and semantic features jointly predict first-time field co-occurrences across a large, multi-domain taxonomy?
    \item Which modeling choices, such as temporal GNN architectures, and neighbor-sampling strategies enhance robustness under increasingly challenging evaluation settings?
    \item Which factors, such as the length of training history or semantic features, most strongly influence predictability on FOS?
\end{enumerate}

To evaluate these questions, we benchmark several state-of-the-art temporal Graph Neural Network (GNN) models on FOS, develop a reproducible evaluation pipeline with multiple negative-sampling regimes, and perform comprehensive ablation studies that probe the contributions of temporal, relational, and semantic signals. Beyond benchmarking, we demonstrate how model outputs can surface promising field pairings, reveal latent collaboration opportunities, and inform strategic research investment decisions. Furthermore, we show that FOS provides a strong temporal-graph benchmark, serving not only the science-forecasting task but also advancing the development and evaluation of temporal GNN models.

Our key contributions are summarized as follows:

\begin{enumerate}
    \item \textbf{FOS Benchmark.} A large-scale temporal graph that captures field co-occurrences among 65,027 research fields across 19 scientific domains, with timestamped edges and semantic node features.
    \item \textbf{Baselines and Evaluation.} A robust and reproducible benchmarking pipeline that evaluates multiple state-of-the-art temporal GNNs under progressively challenging negative-sampling regimes.
    \item \textbf{Ablation Analysis.} Systematic ablation studies that isolate the roles of temporal history, node semantics, and relational structure in predicting novel links.
    \item \textbf{Practical Utility.} Demonstration of how model outputs on FOS can identify emerging interdisciplinary field pairings and guide strategic investment decisions in nascent scientific domains.
\end{enumerate}

\begin{figure*}[t]
    \centering
    \includegraphics[width=\textwidth, height=0.60\textheight, keepaspectratio]{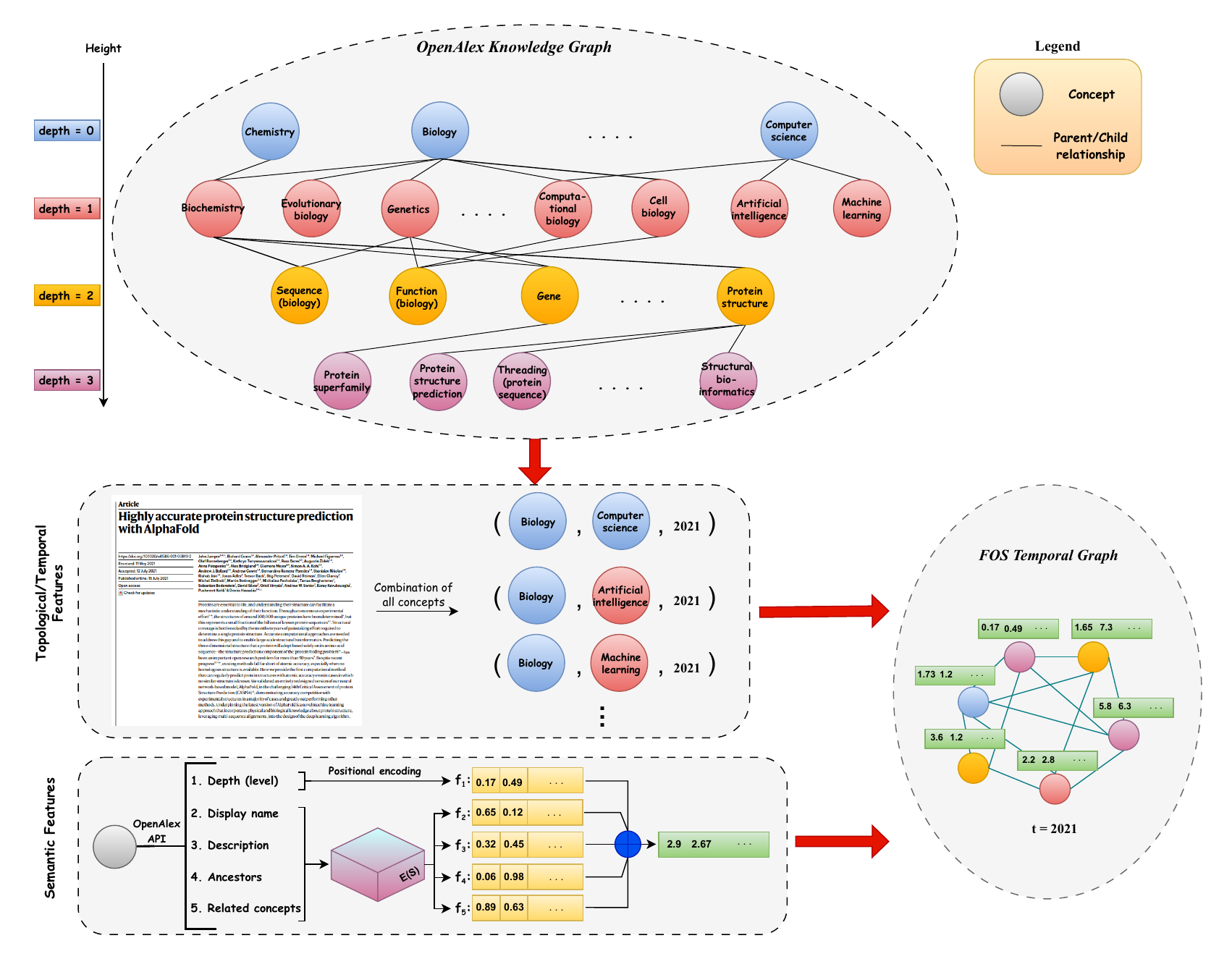}
    \caption{\textbf{Construction of the FOS Temporal Benchmark.} Papers are mapped to concept nodes via the OpenAlex knowledge-graph hierarchy. Co-occurring concepts within a paper form concept pairs that define topological features, while each paper's publication year provides the temporal signal. Each concept node is represented as a semantic vector using an embedding model. These node embeddings, together with the topological and temporal descriptors, are aggregated into annual co-occurrence graphs that collectively constitute the FOS temporal benchmark.}
    \label{fig:main-figs}
\end{figure*}

\section{Related Work}

\subsection{Forecasting Scientific Innovation}

The emergence of new research topics and interdisciplinary collaborations, key concerns of the “science of science” literature, has gained increasing attention within both scientometrics and computational research communities~\cite{fortunato2018science}. Early forecasting efforts predominantly relied on citation trajectories, topic models, and co-word analyses to identify emerging or promising research areas~\cite{borner2018forecasting, salatino2018augur, Schaefermeier2021}. While these methods offered valuable retrospective insights, they were limited in their predictive capacity, as they could not fully capture the dynamic and interdependent evolution of scientific domains. 

More recent studies have reframed scientific forecasting as a temporal knowledge-graph prediction problem, combining temporal co-occurrence patterns, citation networks, and semantic embeddings to model how research concepts evolve over time. For instance, Krenn \textit{et al.}~\cite{krenn2023forecasting} introduce a benchmark based on a dynamic semantic network of roughly 64,000 AI-related concepts, challenging models to forecast novel conceptual links that reflect emerging scientific advances. Similarly, Gu and Krenn~\cite{gu2025forecasting} build dynamic knowledge graphs to predict high-impact research topics, emphasizing how structural changes within scientific ecosystems can provide predictive signals for innovation.

A broader and more heterogeneous approach to forecasting has also been explored. For example, Salatino \textit{et al.}~\cite{salatino2018augur} model the diachronic relationships between research areas using the Adaptive Community Perception Model (ACPM), identifying debutant topics before formal recognition. Expanding on this, Ofer \textit{et al.}~\cite{ofer2024s} combine publication time series, pretrained language models, and patent data to predict the popularity of 125 scientific concepts over a five-year horizon, showcasing the potential of multimodal trend modeling. Likewise, Zhang \textit{et al.}~\cite{zhang2016topic} use text mining, clustering, and technology roadmapping to anticipate domain shifts, using big data research as a case study. 

Collectively, these studies represent a shift from static bibliometric analysis toward dynamic, data-driven, and semantically enriched forecasting frameworks, setting the stage for models that not only observe but also explain the mechanisms behind scientific innovation. Building on this trajectory, our work contributes a large-scale, multi-domain temporal benchmark that systematically advances the modeling of long-horizon, inductive link formation in scientific knowledge graphs.

\subsection{Temporal Knowledge Graphs and Link Prediction}
A substantial body of research on temporal and dynamic graph learning has focused on predicting future relationships, such as link formation or state transitions, based on time-stamped interactions or evolving graph snapshots~\cite{JODIE, trivedi2019dyrep, TGAT, poursafaei2022towards, cong2023we, DyGformer}. Early models, such as JODIE~\cite{JODIE}, employed coupled recurrent neural networks to learn joint embedding trajectories for entities over time, capturing interaction dynamics. Building on this foundation, more recent attention-based architectures, like TGAT~\cite{TGAT}, introduced functional time encodings within graph self-attention layers to better model complex temporal and topological dependencies. Further advances in transformer-based models, exemplified by DyGFormer~\cite{DyGformer}, have introduced novel mechanisms such as neighbor co-occurrence encoding and patch-based history modeling, enabling better handling of long-range temporal dependencies in dynamic link prediction and node classification tasks.

Also, these studies~\cite{zheng2025survey, poursafaei2022towards, huang2023temporal, gastinger2024tgb} have formalized task taxonomies for dynamic GNNs, explored methods for managing temporal and structural evolution, and highlighted key challenges in scalability, inductive generalization, and the adaptation of graph structures. Building on these insights, we adopt key modeling and evaluation practices, particularly inductive and historical evaluation regimes~\cite{poursafaei2022towards}, while tailoring them to the domain of scientific innovation forecasting. Our work connects methodological advances in temporal GNNs with the broader goal of modeling and predicting the evolution of scientific knowledge.

\subsection{Representations of Science and Benchmarks}
A complementary line of research models the evolution of science as a dynamic heterogeneous graph, comprising entities such as authors, papers, and research concepts, and develops specialized datasets to support forecasting tasks. The Science4Cast benchmark~\cite{krenn2023forecasting}, focused on the AI domain, exemplifies this approach by constructing an evolving semantic concept network and evaluating various statistical and machine learning models. Large-scale bibliographic resources such as OpenAlex~\cite{priem2022openalex} further support this effort by providing broad coverage and hierarchical taxonomies of scientific concepts, enabling the construction of fine-grained, time-aware co-occurrence graphs. Despite these advances, existing datasets often remain confined to a single scientific domain and lack the temporal granularity or standardized evaluation protocols required to assess inductive link formation. As a result, the ability to model and forecast interdisciplinary emergence across the entire scientific landscape remains underdeveloped.

\section{Method} \label{proposed}

In this section, we present FOS, a large-scale temporal-graph benchmark designed to model and forecast the evolution of scientific progress. FOS utilizes the OpenAlex knowledge graph, which consists of a hierarchical concept taxonomy, along with publication records, to generate a yearly sequence of graphs that spans the historical record, each representing cross-field links for that year. The benchmark is designed to evaluate methods that (i) capture long-range temporal dynamics, (ii) identify emergent interdisciplinary connections, and (iii) perform robustly under conditions of extreme sparsity and class imbalance. Below, we describe how the OpenAlex knowledge graph is utilized in FOS, formalize the temporal graph representation, and state the forecasting objective as a temporal link-prediction task. Figure~\ref{fig:main-figs} illustrates the pipeline used to create the FOS benchmark.

\subsection{OpenAlex Knowledge Graph}

We construct the FOS dataset using two primary artifacts from the OpenAlex knowledge graph: the Concepts taxonomy and the publication corpus. The Concepts taxonomy provides a hierarchical classification of research topics, consisting of 19 root domains and 5 descendant layers, resulting in a fixed vocabulary of $N = 65{,}027$ distinct fields. Each publication in OpenAlex is tagged with one or more fields of study; when a fine-grained subfield (child) is assigned, all ancestor fields up to the root are implicitly considered assigned as well. Although hierarchical propagation produces obvious parent–child relationships in the raw counts, our modeling emphasis is placed on co-occurrence at the descendant (subfield) level to highlight fine-grained interdisciplinary interactions. For FOS, we use the complete historical span of available publications (years 1827--2024). Additional details on 19 general domains are provided in the Appendix.


\begin{table*}[t]
\centering
\renewcommand{\arraystretch}{1.2} 
\setlength{\tabcolsep}{12pt} 
\begin{tabular}{l|cc|cc|cc}
\toprule
\multirow{2}{*}{\textbf{Model}} & \multicolumn{2}{c|}{\textbf{random}} & \multicolumn{2}{c|}{\textbf{inductive}} & \multicolumn{2}{c}{\textbf{historical}} \\
& \textbf{AP} & \textbf{AUC-ROC} & \textbf{AP} & \textbf{AUC-ROC} & \textbf{AP} & \textbf{AUC-ROC} \\
\midrule
JODIE        & 0.7572 & 0.7594 & 0.6485 & 0.6497 & 0.5671 & 0.5655 \\
DyRep        & 0.8424 & 0.8596 & 0.7193 & 0.7554 & 0.6016 & 0.6315 \\
EdgeBank     & 0.7697 & 0.8412 & 0.6453 & 0.6442 & 0.4852 & 0.4689 \\
GraphMixer   & 0.8813 & 0.8975 & 0.7481 & 0.7682 & \textbf{0.6221} & 0.6367 \\
TGAT         & 0.9105 & 0.9220 & \textbf{0.7565} & \textbf{0.7719} & 0.6030 & 0.6162 \\
DyGFormer    & \textbf{0.9233} & \textbf{0.9338} & 0.7108 & 0.7385 & 0.6066 & \textbf{0.6415} \\
\bottomrule
\end{tabular}
\vspace{2mm}
\caption{\textbf{Main Benchmark Results on FOS\textsubscript{art\&business}.} AP and AUC-ROC scores are reported under three negative-sampling regimes: \emph{random}, \emph{inductive}, and \emph{historical}. Each regime captures a distinct evaluation setting, reflecting varying levels of temporal and structural novelty. The highest score in each column is highlighted in bold.}
\label{tab:main_results}
\end{table*}

\subsection{FOS Benchmark}

\label{FOS}

We model the interactions between fields of study as a temporal graph that consists of a sequence of annual snapshots over a fixed set of vertices. The set $V = \{v_1, \dots, v_N\}$ represents the vocabulary of $N$ fields and serves as the nodes in this graph. The time horizon is defined by the discrete index set $\mathcal{T}$, which includes the years from $t_{\min}$ to $t_{\max}$, where $t_{\min} = 1827$ and $t_{\max} = 2024$. For each year $t \in \mathcal{T}$, the co-occurrence count between subfields $v_m$ and $v_n$, both of which are elements in $V$, is given by $w_t(v_m,v_n)$, which represents the size of the set of papers $P$ such that each paper $p$ is published in year $t$ and is associated with both subfields $v_m$ and $v_n$.

These counts form a weighted adjacency matrix $W_t \in \mathbb{N}^{N \times N}$, where $(W_t)_{mn} = w_t(v_m,v_n)$. To focus solely on the presence of relations, we further construct a binary adjacency matrix, defined as:
\begin{equation}
A_t^{mn} = \begin{cases}
   \mathbf{1} & w_t(v_m,v_n) > 0,\\
   \mathbf{0} & o.w.
\end{cases}
\end{equation}

From this matrix, we define the undirected edge set $E_t = \{(v_m,v_n) \in V \times V : A_t^{mn} = 1\}$. Thus, the graph snapshot for year $t$ is $G_t = (V, E_t)$. Furthermore, to capture historical activity, we compute cumulative matrices:
\begin{equation}
 A_{\le t}^{mn} = \begin{cases}
   \mathbf{1} & \sum_{s\le t}{A_s^{mn}} > 0,\\
   \mathbf{0} & o.w.
\end{cases}
\end{equation}

The first observation time for a pair of subfields $\tau(v_m, v_n)$ is defined as the earliest time step \( t \) in which the pair co-occurs in the dataset, that is, when \( A_t^{mn} = 1 \). If no co-occurrence is ever observed for the pair, its observation time is set to infinity ($\tau(v_m, v_n) = \infty$). We maintain a fixed vertex set $V$ across all years, ensuring nodes remain in the graph even during periods of isolation. This approach provides a consistent domain for temporal modeling and facilitates analysis of phenomena such as long-term activation, dormancy, and re-emergence of subfields~\cite{mertzios2023computing, froese2020comparing}.

In addition to the topological and temporal signals, each node $v_i \in V$ is enriched with a semantic feature vector that captures its hierarchical, lexical, and contextual properties within the ontology. Specifically, we employ the \textbf{AllenAI} model ~\cite{cohan2020specter}, a language model pretrained on a large corpus of scientific documents (including titles and abstracts), to generate embeddings that accurately capture document-level relatedness, making them well-suited for our task.

Formally, let:

\begin{equation}
\mathrm{E} : \mathcal{S} \;\to\; \mathbb{R}^d
\end{equation}
denote the mapping that transforms any text string \(s \in \mathcal{S}\) to a \(d\)-dimensional vector via the AllenAI model. For each node \(v_i\), we compute a set of five feature embeddings:

\begin{equation}
\mathcal{F}(v_i) = \{\, f_j(v_i) \mid f_j(v_i) \in \mathbb{R}^d,\; j=1,\dots,5 \,\}.
\end{equation}
where \(d = 768\). These features correspond to distinct textual aspects. Below we discuss the details of each feature:

\noindent\textbf{Hierarchical Positional Encoding ($f_1$).}  
Let $\ell(v_i)\in\mathbb{N}$ denote the depth or level of a node $v_i$ in the ontology hierarchy; this value is extracted from the OpenAlex concept tree. We embed this level into a fixed‐dimensional sinusoidal positional encoding defined by:

\begin{equation}
\begin{aligned}
(f_1(v_i))_{2j} &= \sin\!\left(\frac{\ell(v_i)}{10000^{\frac{2j}{d}}}\right),\\
(f_1(v_i))_{2j+1} &= \cos\!\left(\frac{\ell(v_i)}{10000^{\frac{2j}{d}}}\right),
\end{aligned}
\quad
j = 0, 1, \dots, \frac{d}{2} - 1.
\end{equation}

Here, the resulting vector $f_1(v_i)\in \mathbb{R}^d$ injects structural positional information derived from the node’s hierarchical level.

\noindent \textbf{Display Name Embedding (\(f_2\)).}  
Let \(\mathrm{name}(v_i)\in \mathcal{S}\) be the canonical short label (display name) of node \(v_i\). We then define
\begin{equation}
f_2(v_i) = \mathrm{E}\bigl(\mathrm{name}(v_i)\bigr).
\end{equation}

\noindent \textbf{Textual Description Embedding (\(f_3\)).}  
Let \(\mathrm{desc}(v_i)\in \mathcal{ S}\) be longer explanatory description for node $v_i$. We define $f_3$ as:
\begin{equation}
f_3(v_i) = \mathrm{E}\bigl(\mathrm{desc}(v_i)\bigr).
\end{equation}

\noindent \textbf{Ancestor Aggregation (\(f_4\)).}  
Let \(\mathrm{Anc}(v_i) = \{a_1, \dots, a_m\}\subseteq \mathcal{S}\) be the set of ancestor labels of node \(v_i\). Then
\begin{equation}
f_4(v_i) = 
\begin{cases}
\displaystyle \frac{1}{m} \sum_{j=1}^m \mathrm{E}(a_j), & m > 0,\\
\mathbf{0}, & m = 0
\end{cases}
\end{equation}
where \(\mathbf{0} \in \mathbb{R}^d\) is the zero vector.

\noindent \textbf{Related-Concept Aggregation (\(f_5\)).}  
Let \(\{r_1, \dots, r_k\}\subseteq \mathcal{S}\) be the set of related concept texts for \(v_i\). Then
\begin{equation}
f_5(v_i) = 
\begin{cases}
\displaystyle \frac{1}{k} \sum_{j=1}^k \mathrm{E}(r_j), & k > 0,\\
\mathbf{0}, & k = 0.
\end{cases}
\end{equation}

The final semantic vector for node $v_i$ is the elementwise sum:
\begin{equation}
e_{v_i} = \sum_{j=1}^5{f_j(v_i)}  \in \mathbb{R}^d.
\end{equation}

This compact representation can be supplied to temporal link-prediction models,  concatenated with temporal and topological descriptors, to improve forecasting of new co-occurrences by utilizing both structural and semantic cues.

\subsection{Forecasting Future of Science via Link Prediction}

We formulate the forecasting of scientific progress as a temporal link prediction problem over the FOS benchmark. Given a sequence of observed yearly graphs $\{G_s\}_{s = t_{\min}}^{t}$
, where each snapshot $G_s=(V,E_s)$ encodes co-occurrences between subfields, the objective is to infer the likelihood of future collaborations or conceptual intersections between any two nodes $(u,v)\in V\times V$ at time $t' > t$. Formally, the task is to estimate the conditional probability
\begin{equation}
P\bigl((u,v)\in E_{t'} \mid \{G_s\}_{s=t{\min}}^{t}\bigr),
\quad t' > t.
\end{equation}

We distinguish between two categories of predictive targets: \textbf{Recurrent links}, corresponding to pairs $(u,v)$ for which a prior co-occurrence exists, i.e., $\tau(u,v)\leq t$; and \textbf{Emerging links}, defined by pairs that first appear after time $t$ (i.e., $t<\tau(u,v)$). The latter category is of particular scientific interest, as it captures the first-time convergence of previously unconnected research areas, effectively modeling the emergence of novel scientific directions.

\section{Experiments}\label{experiments}
In this section, we evaluate FOS through three complementary analyses: (i) benchmarking a diverse suite of state-of-the-art temporal graph models, (ii) assessing their performance under multiple realistic evaluation regimes, and (iii) conducting targeted ablations to quantify the effects of temporal horizons in training data and the contribution of semantic features to forecasting performance.

\subsection{Model Selection}
To benchmark FOS and establish strong empirical baselines, we evaluate a representative set of
state-of-the-art temporal graph models. Specifically, we study JODIE (event-driven)~\cite{JODIE}, DyRep (attentive recurrent)~\cite{trivedi2019dyrep}, TGAT (transformer-based)~\cite{TGAT}, EdgeBank (memory-based)~\cite{poursafaei2022towards}, GraphMixer (MLP-based)~\cite{cong2023we}, and DyGFormer (hybrid)~\cite{DyGformer}. Collectively, these models span a broad spectrum of inductive biases, from temporal event modeling and recurrent updates to attention mechanisms, memory aggregation, and channel mixing, enabling a systematic assessment of which architectural principles are most effective for FOS’s long-horizon, sparse, and highly imbalanced link prediction task.

\subsection{Evaluation Setup and Implementation Details}

We evaluate the selected baselines under three negative-sampling settings, random, historical, and inductive, combined with three temporal neighborhood sampling strategies: uniform, recent, and time-interval-aware, following established protocols in prior works~\cite{poursafaei2022towards, DyGformer}. Performance is measured using Average Precision (AP) and AUC-ROC, computed per test batch and averaged across test years to account for temporal variation. Detailed definitions of these metrics, as well as the procedures for negative and neighborhood sampling, are provided in the Appendix.

Experiments are conducted on a representative FOS subset encompassing the Art and Business domains and their descendants, comprising 3,238 nodes and 3,472,315 edges (FOS\textsubscript{art\&business}). The temporal split includes training from 2002–2017, validation from 2018–2021, and testing from 2022–2024. For all baselines, the configuration consists of two GNN layers with two attention heads, a batch size of 300, 20 sampled neighbors, a walk length of 1 with 8 walk heads, a time-feature dimension of 100, and a positional embedding dimension of 172. Training is performed using the Adam optimizer with a learning rate of $1\times10^{-4}$, a dropout rate of 0.1, over 30 epochs with early stopping (patience of 20).

\begin{figure*}
    \centering
    \includegraphics[width=1\linewidth]{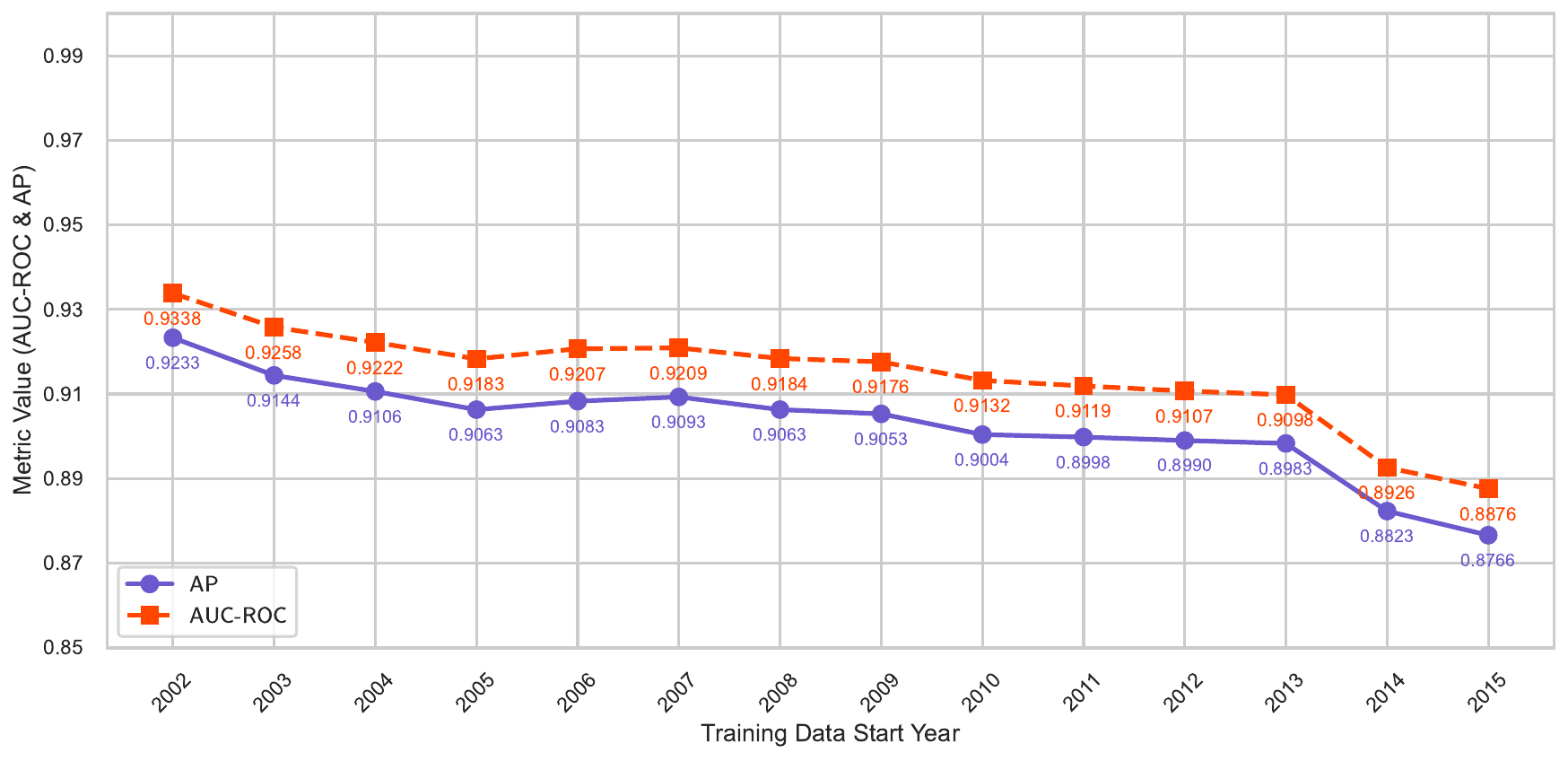}
    \caption{\textbf{Impact of Historical Data Span on Model Performance.} This plot illustrates AP and AUC-ROC scores, averaged across test years (2022-2024), for the DyGFormer model trained on varying historical data windows starting from 2002 to 2015 (with a fixed end year). Extending the historical span improves predictive performance, though with diminishing marginal gains as the starting year recedes further into the past.}
    \label{fig:temporal_ablation}
\end{figure*}

\subsection{Main Results}

Table~\ref{tab:main_results} summarizes the performance of the baseline models on the proposed FOS\textsubscript{art\&business} benchmark. We report Average Precision (AP) and AUC-ROC across three negative-sampling settings: random, inductive, and historical.

In the random setting, DyGFormer achieves the highest scores, with an AP of \textbf{0.9233} and an AUC-ROC of \textbf{0.9338}, surpassing the next best model, TGAT, by absolute margins of 0.0128 AP (1.41\%) and 0.0118 AUC-ROC (1.28\%). In the inductive setting, TGAT outperforms the other models with an AP of 0.7565 and AUC-ROC of 0.7719, while DyGFormer records an AP of 0.7108 and AUC-ROC of 0.7385. The AP gap of 0.0457 (6.04\%) highlights that architectures which perform well under random sampling do not necessarily generalize best to inductive scenarios. In the historical setting, GraphMixer leads in AP with a score of 0.6221, while DyGFormer achieves the highest AUC-ROC at 0.6415. The difference in AP between GraphMixer and DyGFormer is 0.0155 (2.49\%), while DyGFormer’s AUC-ROC advantage over GraphMixer is 0.0048 (0.75\%). This pattern, stronger AUC-ROC but not always superior AP, suggests that models exhibit varying behaviors in ranking versus discrimination when forecasting temporally evolving links.

Two key takeaways emerge from these results. First, our benchmark reveals considerable variation in model performance across the different evaluation settings: accuracy typically declines as the evaluation becomes more challenging (random → inductive → historical). For example, DyGFormer’s AP drops from 0.9233 in the random setting to 0.7108 in the inductive setting, and further to 0.6066 in the historical setting. This highlights the added difficulty of forecasting in historical and inductive settings. Second, no single model consistently outperforms across all regimes; different architectures exhibit complementary strengths. These findings validate FOS as a challenging and discriminative benchmark, one that effectively differentiates model performance and motivates future research into architectures and training strategies for temporal, sparse, and highly imbalanced scientific co-occurrence prediction.

\subsection{Ablation Study}

We conduct two complementary ablations to elucidate the factors that drive model performance on FOS:
(i) a \textit{temporal analysis}, which quantifies the contribution of historical training data to predictive accuracy, and
(ii) a \textit{node feature analysis}, which assesses the relative importance of each semantic feature described in the previous section.
All ablations use the random negative-sampling setting and report Average Precision (AP) and AUC-ROC, averaged over the test years (2022–2024) for consistency with the main evaluation.

\subsubsection{Temporal Analysis. }
\label{sec:ablation-temporal}

To isolate the influence of training history, we fix the model architecture to DyGFormer and vary the start year of the training period. In the main experiments, the canonical interval is 2002–2017. Here, we progressively truncate the training window by advancing the start year one year at a time (2002, 2003, 2004, $\dots$), while keeping the end year fixed. For each start year, DyGFormer is retrained with identical hyperparameters, ensuring that any observed performance changes stem solely from the reduction in historical context.

Figure~\ref{fig:temporal_ablation} plots AP and AUC-ROC as functions of the training start year. Both metrics exhibit a clear monotonic decline as the training horizon shortens, with only minor fluctuations between 2005–2007. Models trained on longer historical windows achieve consistently higher performance, confirming that long-range co-occurrence history provides valuable contextual information for forecasting novel field associations in FOS. The largest marginal gains arise from including the earliest years of the dataset, while the benefit tapers as increasingly distant history is added, suggesting diminishing returns from very old temporal information. These findings underscore the cumulative, path-dependent nature of interdisciplinary evolution and highlight the importance of extended temporal context for robust forecasting.

\subsubsection{Node Feature Analysis. }
\label{sec:ablation-features}

We evaluate the contribution of each semantic feature through an ablation study using the DyGFormer model. As a reference, the model is first trained with the full set of node features (“full features”). To estimate the marginal importance of a specific feature $f_j$, we retrain DyGFormer while omitting $f_j$, keeping all other inputs, architecture components, and hyperparameters identical. Table~\ref{tab:ablation_features} reports the resulting AP and AUC-ROC scores for each configuration.

\begin{table}[ht]
\centering
\begin{tabular}{@{}>{\raggedright\arraybackslash}p{4cm} c c@{}}
\toprule
\textbf{Model Configuration} & \textbf{AP} & \textbf{AUC-ROC} \\
\midrule
DyGFormer (full features) & 0.9233 & 0.9338 \\
\hspace{3mm}w/o related & 0.9157 & 0.9273 \\
\hspace{3mm}w/o name & 0.9137 & 0.9252 \\
\hspace{3mm}w/o ancestor & 0.9131 & 0.9253 \\
\hspace{3mm}w/o level & 0.9122 & 0.9242 \\
\hspace{3mm}w/o desc & 0.8967 & 0.9027 \\
\bottomrule
\end{tabular}
\caption{\textbf{Ablation Results on DyGFormer:} Impact of omitting individual semantic node features on AP and AUC-ROC (averaged over test years 2022–2024). Each row reports performance when the indicated feature is removed.}
\label{tab:ablation_features}
\end{table}

The results indicate that removing the \texttt{desc} (description embedding) feature produces the largest decline in performance (\textit{–0.0266 AP}, \textit{–0.0311 AUC-ROC}), demonstrating that rich descriptive text provides the most informative single semantic signal for forecasting novel co-occurrences. The remaining features—\texttt{related}, \texttt{name}, \texttt{ancestor}, and \texttt{level}—also contribute positively, though to a lesser extent. Excluding any one of these features typically reduces AP by 0.007–0.011 and yields a similar decrease in AUC-ROC. These consistent yet moderate declines suggest that complementary structural and lexical cues enhance forecasting performance, but that long-form textual semantics remain the dominant predictive factor in DyGFormer’s representation of field-level dynamics.

\begin{table*}[t!]
\centering
\footnotesize
\begin{tabular}{>{\raggedright\arraybackslash}p{8.75cm} >{\centering\arraybackslash}p{1.75cm} >{\centering\arraybackslash}p{1.5cm} >{\raggedright\arraybackslash}p{5cm}}
\toprule
\arrayrulecolor{black}
Title & Publication Date & Prediction Score & Concepts \\
\midrule
\arrayrulecolor{gray!30}
Research on school-enterprise collaborative lighting innovation design based on social psychology and Kansei engineering: Taking the collaboration between central academy of fine arts and Librite as an example \href{https://openalex.org/W4414026208}{W4414026208} & 2025-08-27 & 0.982 & {\sethlcolor{red!20}\hl{Kansei engineering}} - {\sethlcolor{yellow!20}\hl{The arts}} \\
\midrule
The Role of Artificial Intelligence in Enhancing Decision-making and Efficiency in Mergers and Acquisitions: A Case Study Approach within the U.S. Capital Market. \href{https://openalex.org/W4410462891}{W4410462891} & 2025-05-09 & 0.979 & {\sethlcolor{red!20}\hl{Market intelligence}} - {\sethlcolor{yellow!20}\hl{Capital (architecture)}} \\
\midrule
A Framework for Artificial Intelligence Driven Process and Service Innovation in Food and Beverage Departments of Premium Hotels in National Capital Region \href{https://openalex.org/W4410871899}{W4410871899} & 2025-05-29 & 0.979 & {\sethlcolor{red!20}\hl{Market intelligence}} - {\sethlcolor{yellow!20}\hl{Capital (architecture)}} \\
\midrule
Under the Background of “Internet+”, Cultural Products Promote the Inheritance and Innovation of Traditional Culture—Taking the Consumption Boom of “Guochao” Products as an Example \href{https://openalex.org/W4409703909}{W4409703909} & 2025-01-01 & 0.964 & {\sethlcolor{red!20}\hl{Commerce}} - {\sethlcolor{yellow!20}\hl{Cultural inheritance}} \\
\midrule
A Multi-Level Annotation Model for Fake News Detection: Implementing Kazakh-Russian Corpus via Label Studio \href{https://openalex.org/W4413342587}{W4413342587} & 2025-08-20 & 0.964 & {\sethlcolor{red!20}\hl{Fake news}} - {\sethlcolor{yellow!20}\hl{Studio}} \\
\arrayrulecolor{black}
\bottomrule
\end{tabular}
\caption{\textbf{Examples of Cross-Field Links Predicted by DyGFormer}. Selected high-confidence predictions in the Art and Business subset that were realized for the first time in 2025 publications. Columns list the publication title, date, prediction score, and the corresponding concept pair.}
\label{tab:real-world}
\end{table*}

\section{Discussion}

In this section, we examine the broader implications of the FOS benchmark and situate it within the landscape of existing temporal graph benchmarks. First, we explore its potential real-world applications, emphasizing how FOS can serve as a predictive engine for identifying emerging interdisciplinary connections and fostering data-driven scientific collaboration. Next, we analyze FOS as a competitive temporal-graph benchmark, highlighting its temporal dynamics and structural complexity, and positioning it as a useful tool for evaluating temporal link prediction models.

\subsection{Real-World Applications of FOS}

FOS can serve as a valuable tool for identifying emerging interdisciplinary connections in scientific research. To evaluate its predictive power in real-world scenarios, we simulate a deployment by treating 2025 publications as out-of-sample data. Predictions generated from historical data are compared against actual publications from 2025 within the Art and Business FOS subset.

Table~\ref{tab:real-world} presents high-confidence predictions of DyGFormer where predicted cross-field links were appeared for the first time in 2025 publications. Notable examples include \emph{Kansei engineering \(\leftrightarrow\) The arts}, two independent cases of \emph{Market intelligence \(\leftrightarrow\) Capital (architecture)}, \emph{Cultural inheritance \(\leftrightarrow\) Commerce}, and \emph{Fake news \(\leftrightarrow\) Studio}. These cases demonstrate that the model effectively identifies promising cross-field relationships that are likely to materialize into published research within a short time frame, validating FOS's predictive capacity.

For future work, we envision the development of an intelligent collaboration platform that leverages the predictions of temporal GNN models on FOS to not only identify emerging cross-field connections but also suggest potential research partners. By analyzing researchers' past publications, expertise, and complementary skills, such a platform could automatically identify teams capable of tackling complex interdisciplinary challenges. This would transform predicted connections into actionable collaborations, accelerating breakthroughs in areas that might otherwise remain siloed.

\subsection{FOS as a Competitive Temporal-Graph Benchmark}

We assess FOS as a \emph{competitive} temporal-graph benchmark, one that is sufficiently challenging yet allows meaningful exploitation of temporal structure. To evaluate this, we apply the TEA/TET diagnostic framework~\cite{poursafaei2022towards} to characterize FOS’s temporal edge dynamics relative to established benchmarks.

The TEA/TET diagnostics consist of three key metrics: (i) \emph{Novelty}, which measures the percentage of new edges each year relative to previous years; (ii) \emph{Recurrence}, the fraction of edges in the test period that reappear from the training period; and (iii) \emph{Surprise}, the proportion of edges in the test period that are completely new (never seen during training).

We compute these metrics for FOS\textsubscript{art\&business} and compare them against several established temporal graph datasets. The results, presented in Table~\ref{tab:novelty_recurrence_surprise}, position FOS\textsubscript{art\&business} in an \emph{intermediate regime}: it shows moderate novelty (novelty = 0.19) and surprise (surprise = 0.11), with a significant portion of edges recurring (recurrence = 0.40). This balance indicates that FOS is not trivially reducible to replaying past edges, nor does it consist entirely of novel inductive challenges. 

To confirm the robustness of this intermediate regime, we evaluate a memorization baseline (EdgeBank) and observe that it does not achieve strong performance in our experiments, as shown in Table~\ref{tab:main_results}. This result demonstrates that simply memorizing historical edges is insufficient, further supporting the claim that FOS presents a nontrivial challenge. 

Taken together, the TEA/TET diagnostics and empirical results confirm that FOS is a \emph{competitive} and \emph{realistic} temporal-graph benchmark: it presents a meaningful challenge while retaining exploitable structure, making it an appropriate testbed for advancing dynamic link prediction models.

\begin{table}[ht]
  \centering
  \begin{tabular}{l c c c}
    \toprule
    Dataset & Novelty $\uparrow$ & Recurrence $\downarrow$ & Surprise $\uparrow$ \\
    \midrule
    Wikipedia~\cite{JODIE}        & 0.46 & 0.26 & 0.42 \\
    MOOC~\cite{JODIE}             & 0.75 & 0.02 & 0.79 \\
    LastFM~\cite{JODIE}           & 0.28 & 0.30 & 0.35 \\
    Enron~\cite{shetty2004enron}            & 0.30 & 0.22 & 0.27 \\
    Social Evo.~\cite{madan2011sensing}      & 0.11 & 0.51 & 0.02 \\
    UCI~\cite{panzarasa2009patterns}              & 0.73 & 0.01 & 0.56 \\
    Flights~\cite{schafer2014bringing}          & 0.21 & 0.60 & 0.19 \\
    Can. Parl.~\cite{huang2020laplacian}       & 0.69 & 0.01 & 0.57 \\
    US Legis.~\cite{fowler2006legislative, huang2020laplacian}        & 0.44 & 0.08 & 0.45 \\
    UN Trade~\cite{macdonald2015rethinking}         & 0.07 & 0.87 & 0.04 \\
    UN Vote~\cite{bailey2017estimating}          & 0.03 & 0.93 & 0.01 \\
    Contact~\cite{sapiezynski2019interaction}          & 0.42 & 0.44 & 0.12 \\
    \midrule
    \textbf{FOS\textsubscript{art\&business}} & 0.19 & 0.40 & 0.11 \\
    \bottomrule
  \end{tabular}
  \vspace{2mm}
  \caption{\textbf{TEA/TET Diagnostics Across Temporal-Graph Benchmarks}. Reported metrics include \emph{Novelty}, \emph{Recurrence}, and \emph{Surprise}, which capture complementary aspects of temporal edge dynamics.}
  \label{tab:novelty_recurrence_surprise}
\end{table}

\subsection{Limitations and Future Work}

While \textbf{FOS} represents a significant advance in temporal forecasting of scientific emergence, we acknowledge several important limitations and outline concrete directions for future work.

\paragraph{Domain scope and generalizability.}  
Our empirical evaluation focuses primarily on the Art \& Business subset of FOS for tractability. Hence, the forecasting performance, model behavior, and domain-specific dynamics we observe may not generalize across other scientific domains or to the full 65,027-field taxonomy. In future work, we will extend experiments to additional top-level domains (e.g., Computer Science, Biology) and scale training accordingly to validate cross-domain robustness.

\paragraph{Link-creation scheme and emerging-field bias.}  
In this work, the co-occurrence link-generation method follows prior pipelines~\cite{krenn2023forecasting, gu2025forecasting} but may introduce bias: when one subfield emerges later than another, the timestamp $\tau(v_m, v_n)$ may reflect the newer subfield’s introduction rather than a genuine cross-field pairing. Consequently, some predicted links may capture field births rather than meaningful interdisciplinary interaction. Future work can refine the methodology by filtering or down-weighting such cases and by modelling the “maturation interval” required for a sub-field to integrate into broader research networks.

By addressing these limitations and pursuing the proposed research directions, we believe FOS can evolve into a more broadly applicable, and impactful foundation for modeling the dynamics of science.

\section{Conclusion}

In this work, we introduced the \textbf{FOS} benchmark: a large‐scale, time‐aware co-occurrence graph built from the OpenAlex taxonomy, capturing yearly field–field interactions among 65,027 fields over 19 top‐level domains, enriched with semantic node embeddings and temporal/topological descriptors.  

Our empirical evaluation demonstrates that FOS presents a challenging forecasting regime: while transformer-hybrid models such as DyGFormer achieve the highest performance under the random negative-sampling setting, attention-based and memory-efficient architectures exhibit complementary strengths in inductive and historical regimes. This divergence underlines that no single architecture dominates across forecasting scenarios. Further, our ablation studies reveal that the node “description” embedding (rich textual semantics) is the most informative semantic feature for forecasting novel co-occurrences.

Beyond benchmark outcomes, we show that high‐confidence predictions from temporal GNN models on FOS align with real interdisciplinary publications that emerged shortly after our training period, even within the Art \& Business subset, suggesting that FOS can help surface emergent interdisciplinary linkages in the near future.  

Overall, FOS offers a temporally rich, semantically meaningful, and practically relevant testbed that distinguishes temporal GNN architectures and surfaces open challenges in modeling scientific emergence. We release the dataset, standardized splits, and evaluation scripts to accelerate community adoption. We hope this will enable the broader research community to build more robust forecasting models and ultimately guide scientific discovery toward novel, high-impact interdisciplinary connections.

\bibliography{aaai2026}


\pagebreak

\section{Appendix}
This appendix provides extended descriptions, implementation details, and all resources necessary to ensure full reproducibility of the \textbf{FOS} benchmark. 
Below, we summarize the content of each subsection and what readers can expect to find.

\begin{itemize}
    \item \textbf{\hyperref[app:openalex]{OpenAlex Knowledge Graph Description.}}  
    Explains how the OpenAlex scholarly graph underpins FOS, including the hierarchical taxonomy of 19 top-level domains and 65{,}027 fields. Describes the mapping of concept nodes and the construction of domain--subdomain connections.

    \item \textbf{\hyperref[app:fos_art]{FOS\textsubscript{Art \& Business} Description.}}  
    Characterizes the specific subset used in experiments. Presents descriptive statistics from node-, edge-, graph-, and temporal-pattern perspectives, visualizing activity distributions, edge dynamics, and structural evolution across time.

    \item \textbf{\hyperref[app:models]{Models Description.}}  
    Summarizes the temporal graph neural network architectures evaluated (JODIE, DyRep, TGAT, EdgeBank, GraphMixer, DyGFormer). For each, outlines its temporal encoding strategy, memory mechanism, and inductive learning capacity.

    \item \textbf{\hyperref[app:metrics]{Metrics Description.}}  
    Defines the quantitative evaluation metrics used, Average Precision (AP) and AUC-ROC, and clarifies their computation in temporal link-prediction settings for consistent cross-model comparison.

    \item \textbf{\hyperref[app:negative_sampling]{Negative Sampling Strategies.}}  
    Details the construction of negative samples under three regimes: \emph{Random}, \emph{Historical}, and \emph{Inductive}. Explains how these sampling pools control the level of temporal and structural novelty during evaluation.

    \item \textbf{\hyperref[app:neighbor_sampling]{Neighbor Sampling Strategies.}}  
    Describes the three neighborhood construction techniques, \emph{Uniform}, \emph{Recent}, and \emph{Time-interval-aware}, used to build temporal node contexts. Provides the underlying equations for time-biased sampling probabilities.

    \item \textbf{\hyperref[app:implementation]{Implementation Details.}}  
    Provides complete experiment configurations, including dataset splits, node-feature preprocessing (e.g., PCA reduction), model hyperparameters, optimizer settings, batch formation, and computational setup to enable reproducibility.

    \item \textbf{\hyperref[app:predictions]{List of Predictions.}}  
    Extends the examples discussed in Table~\ref{tab:real-world} by listing high-confidence predicted cross-field links. Demonstrates how these model forecasts correspond to real interdisciplinary publications emerging in subsequent years.

    \item \textbf{\hyperref[app:tea_tet]{TEA and TET Plots.}}  
    Presents diagnostic visualizations, \textit{Temporal Edge Appearance (TEA)} and \textit{Temporal Edge Traffic (TET)}, that capture novelty, recurrence, and surprise dynamics in FOS. These plots validate the temporal realism and structural diversity of the benchmark.
\end{itemize}

\subsection{OpenAlex Knowledge Graph Description}
\label{app:openalex}

The OpenAlex knowledge graph is a large, open scholarly graph that indexes works, authors, venues, institutions, and conceptual topics. In this work we use OpenAlex as the reference scholarly taxonomy and concept hierarchy that underlies our FOS temporal-graph benchmark.

Table~\ref{tab:connected_nodes} reports, for each top-level domain, the number of OpenAlex concept/subfield nodes that are \emph{connected} to that domain within the OpenAlex hierarchical taxonomy. By ``connected'' we mean that the counted node has at least one parent/child path that reaches the corresponding top-level domain node.

Concretely, the counts in Table~\ref{tab:connected_nodes} are computed as follows:
\begin{enumerate}
  \item We treat OpenAlex concept entities as nodes in the concept hierarchy.
  \item A node is considered \emph{connected} to a top-level domain \(D\) if there exists at least one path following parent/child links from the node to \(D\).
  \item Each node is counted at most once per domain (nodes that have paths to multiple top-level domains appear in each relevant domain's count).
\end{enumerate}

For example, in this study we focus on two top-level domains, Business and Art, and include all nodes that are hierarchically connected to them in OpenAlex. This subset comprises 3,266 nodes in total and constitutes the FOS\textsubscript{art\&business} dataset used in our experiments.

\begin{table}[ht]
\centering
\begin{tabular}{l r}
\toprule
\textbf{Domain} & \textbf{Connected Nodes} \\
\midrule
Biology               & 24,780 \\
Medicine              & 17,775 \\
Chemistry             & 14,828 \\
Physics               & 12,341 \\
Computer Science      & 8,202  \\
Engineering           & 6,947  \\
Mathematics           & 6,088  \\
Psychology            & 5,201  \\
Political Science     & 4,433  \\
Geology               & 4,379  \\
Materials Science     & 3,982  \\
Economics             & 3,929  \\
Geography             & 3,070  \\
Philosophy            & 2,968  \\
History               & 2,237  \\
Sociology             & 2,125  \\
Business              & 1,960  \\
Art                   & 1,306  \\
Environmental Science &   484  \\
\bottomrule
\end{tabular}
\vspace{2mm}
\caption{\textbf{Number of Connected OpenAlex Nodes per Top-level Domain.} Counts represent the number of distinct concept or subfield nodes that are hierarchically linked to each top-level domain in the OpenAlex taxonomy.}
\label{tab:connected_nodes}
\end{table}

\subsection{FOS\textsubscript{art\&business} Description}
\label{app:fos_art}

In this subsection, we provide a detailed characterization of our temporal-graph benchmark, highlighting the dynamic structure of the FOS\textsubscript{art\&business} dataset. These characteristics are categorized into node-level, edge-level, graph-level, and temporal-pattern (dynamical) perspectives, each offering a distinct lens on the dataset:

\begin{enumerate}
    \item \textbf{Node-level features} capture local activity patterns for individual nodes, including measures of activity, persistence, connectivity growth, and local structure.
    \item \textbf{Edge-level features} examine pairwise interactions, such as recurrence frequency, lifespan, inter-event timing, and recency of appearances.
    \item \textbf{Graph-level features} offer a holistic view of network topology, encompassing connectivity, component structure, and global reachability.
    \item \textbf{Temporal-pattern features} quantify dynamical processes over time, including edge turnover and temporal mixing of node activities.
\end{enumerate}

\begin{figure*}
    \centering
    \includegraphics[width=1\linewidth]{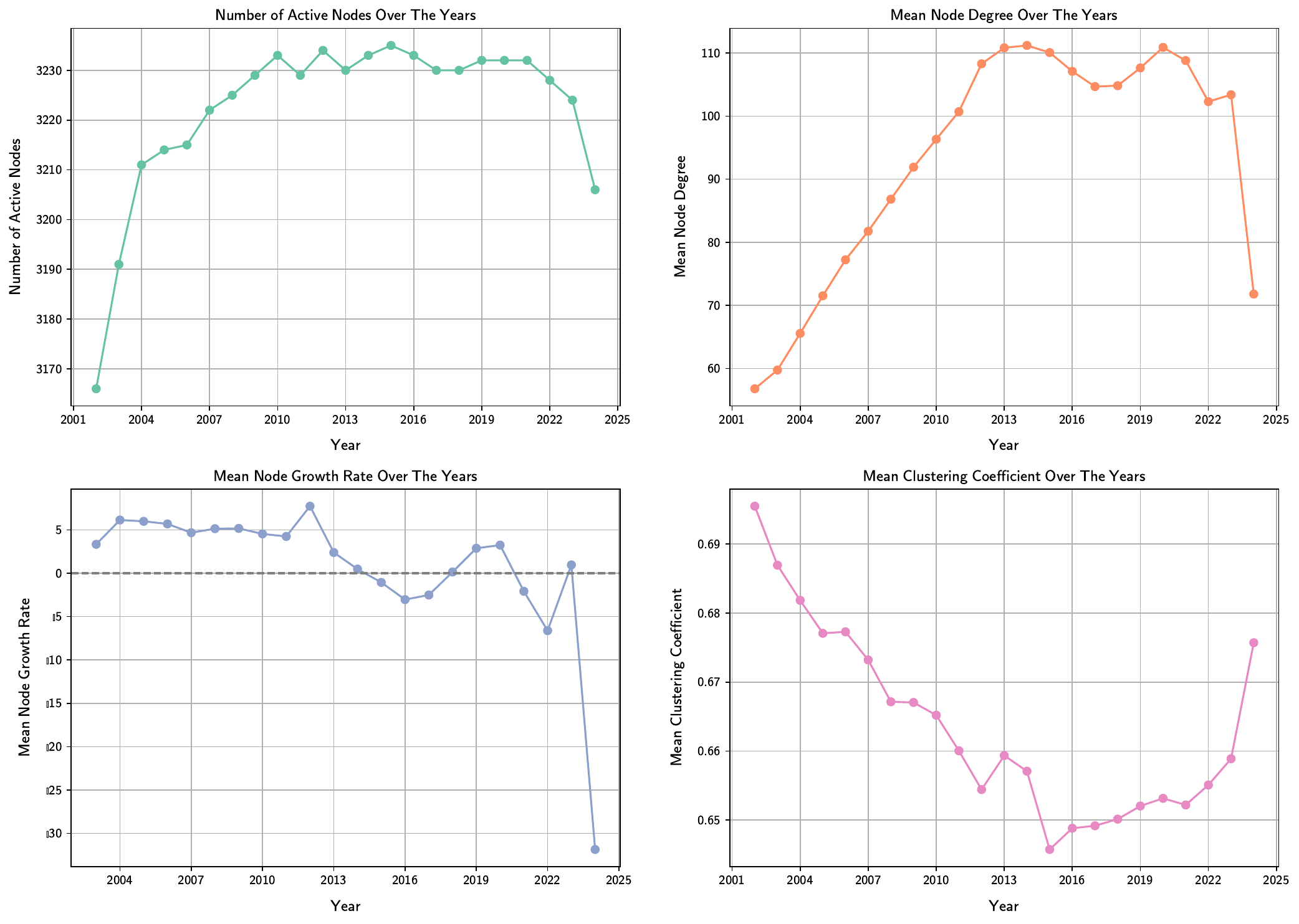}
    \caption{\textbf{FOS\textsubscript{art\&business} Node-Level Statistics (Part 1):} Plots depicting (1) the number of active nodes per year, (2) the mean node degree per year, (3) the mean annual growth rate in node degrees relative to the prior year, and (4) the mean clustering coefficient per year.}
    \label{fig:node_statistics_1}
\end{figure*}

\begin{figure*}
    \centering
    \includegraphics[width=1\linewidth]{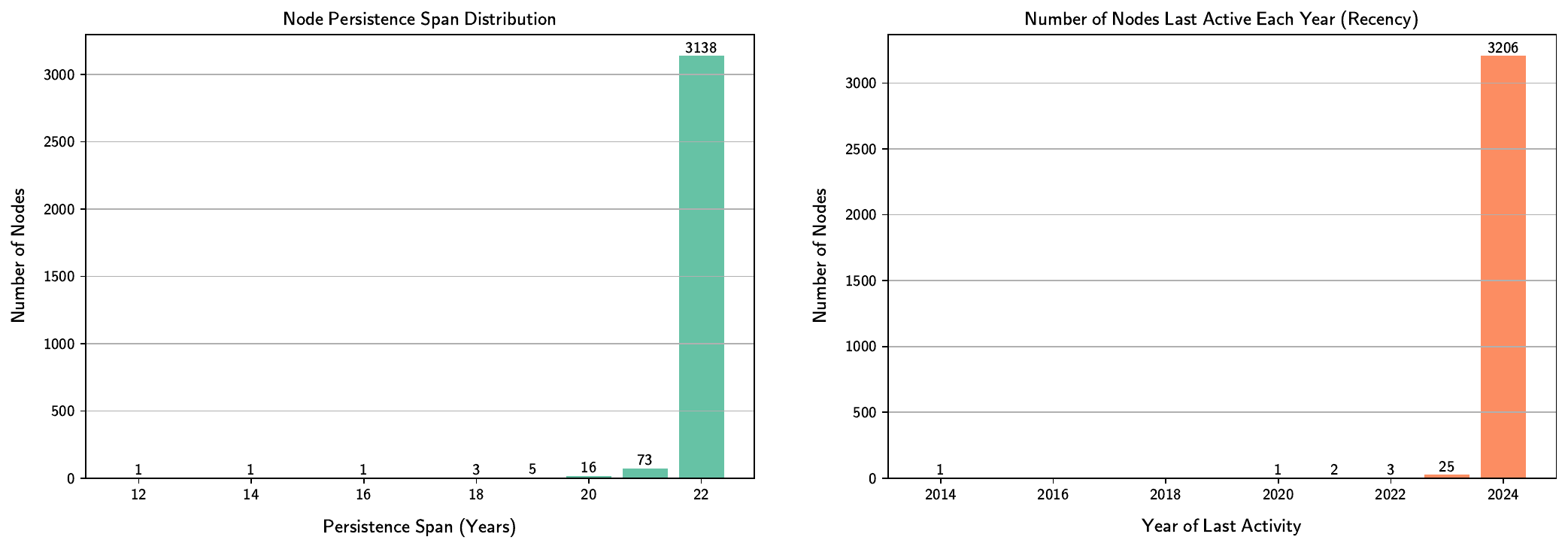}
    \caption{\textbf{FOS\textsubscript{art\&business} Node-Level Statistics (Part 2):} Distributions of (1) node persistence spans (total active years) and (2) the number of nodes by year of last activity (recency).}
    \label{fig:node_statistics_2}
\end{figure*}

\begin{figure*}
    \centering
    \includegraphics[width=1\linewidth]{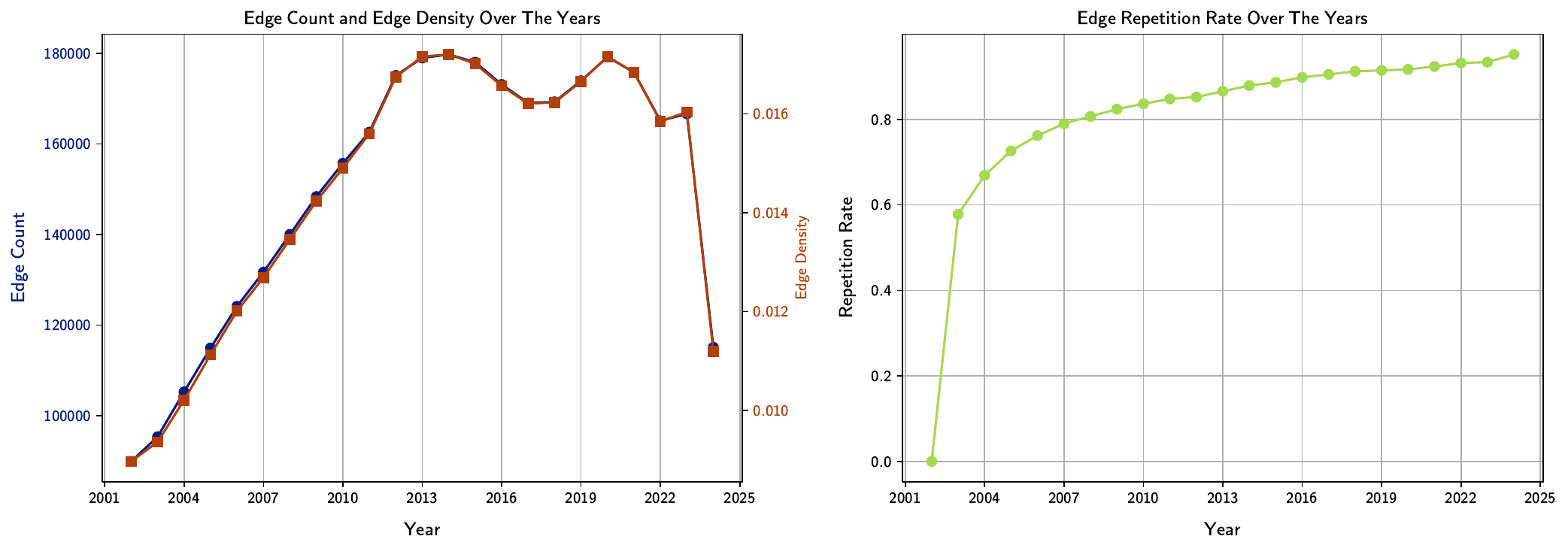}
    \caption{\textbf{FOS\textsubscript{art\&business} Edge-Level Statistics (Part 1):} Plots showing (1) edge count and density over time and (2) the edge repetition rate (fraction of recurring edges) per year.}
    \label{fig:edge_statistics_1}
\end{figure*}

\begin{figure*}
    \centering
    \includegraphics[width=1\linewidth]{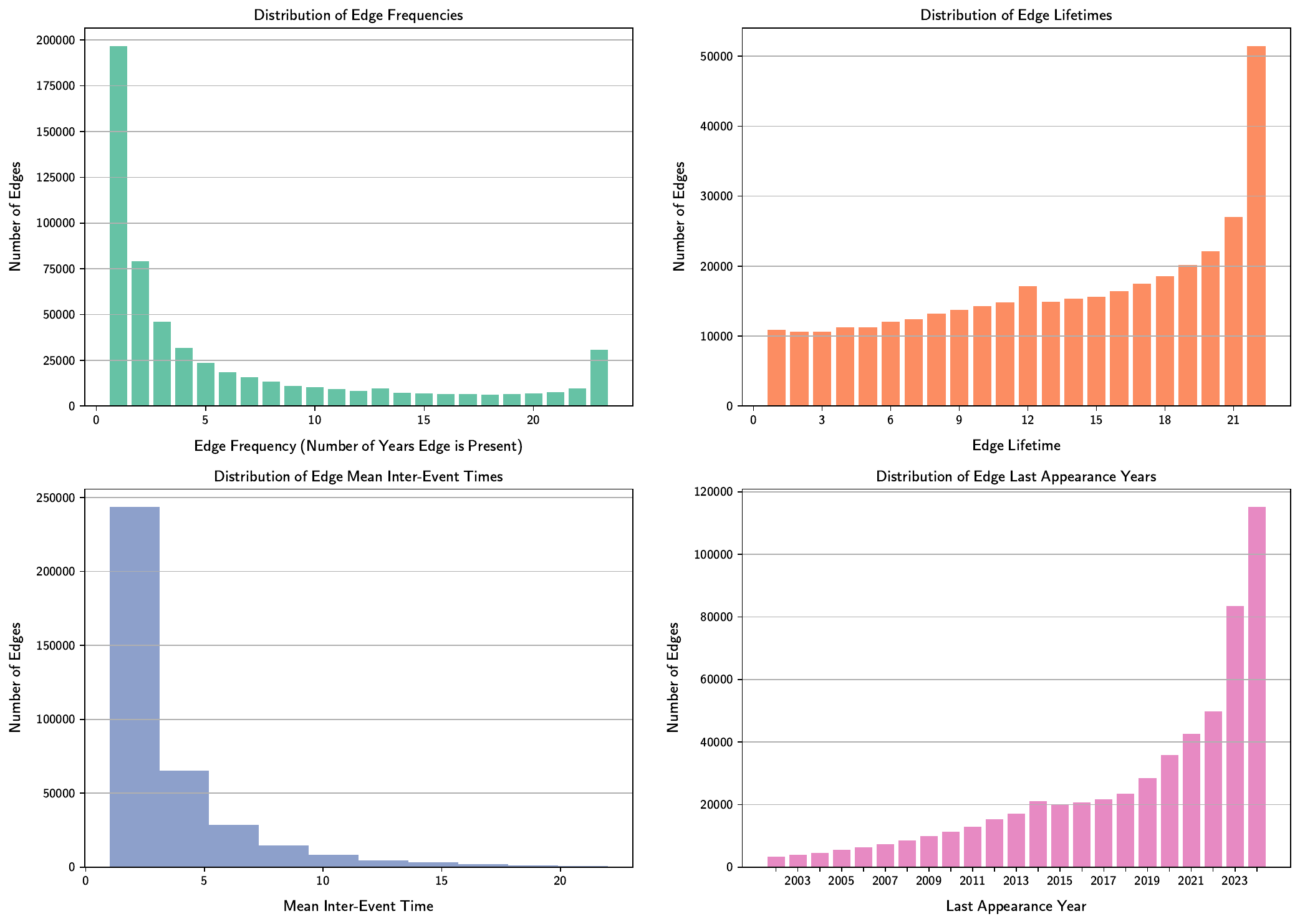}
    \caption{\textbf{FOS\textsubscript{art\&business} Edge-Level Statistics (Part 2):} Distributions of (1) edge frequencies (years present), (2) edge lifetimes (span from first to last appearance), (3) mean inter-event times for edges, and (4) edge last appearance years.}
    \label{fig:edge_statistics_2}
\end{figure*}

\begin{figure*}
    \centering
    \includegraphics[width=1\linewidth]{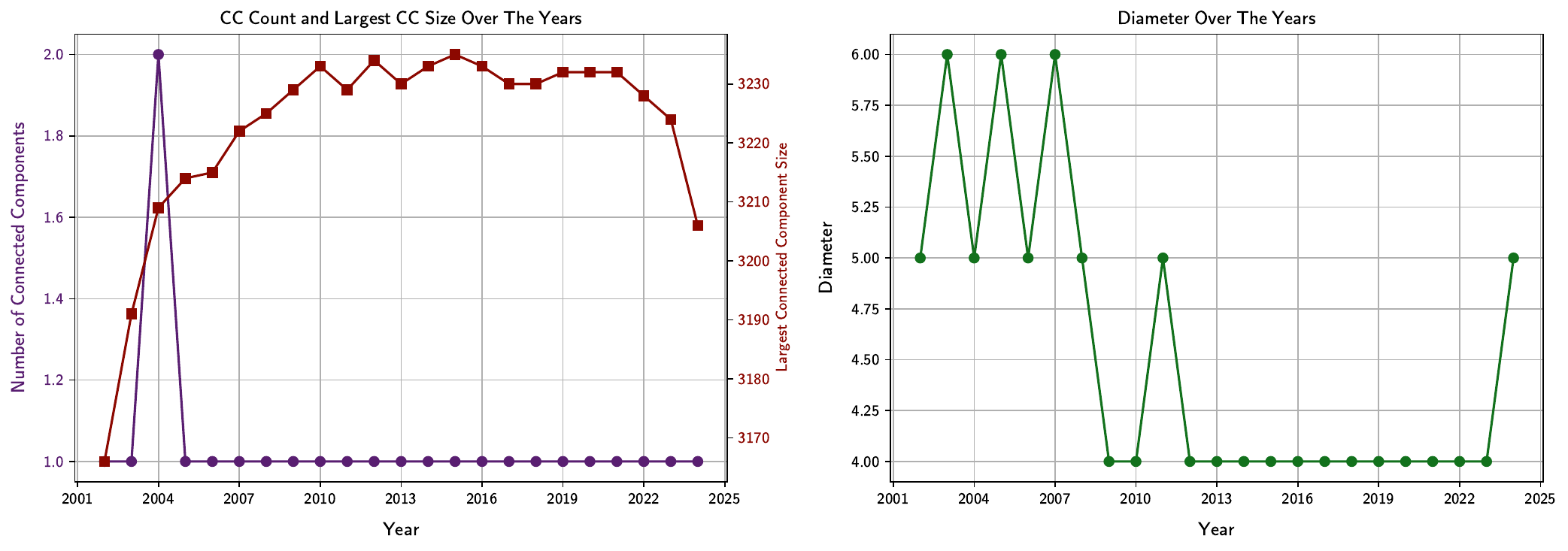}
    \caption{\textbf{FOS\textsubscript{art\&business} Graph-Level Statistics:} Plots illustrating (1) the number of connected components (CCs) and largest CC size over time and (2) the graph diameter per year.}
    \label{fig:graph_statistics}
\end{figure*}

\begin{figure*}
    \centering
    \includegraphics[width=1\linewidth]{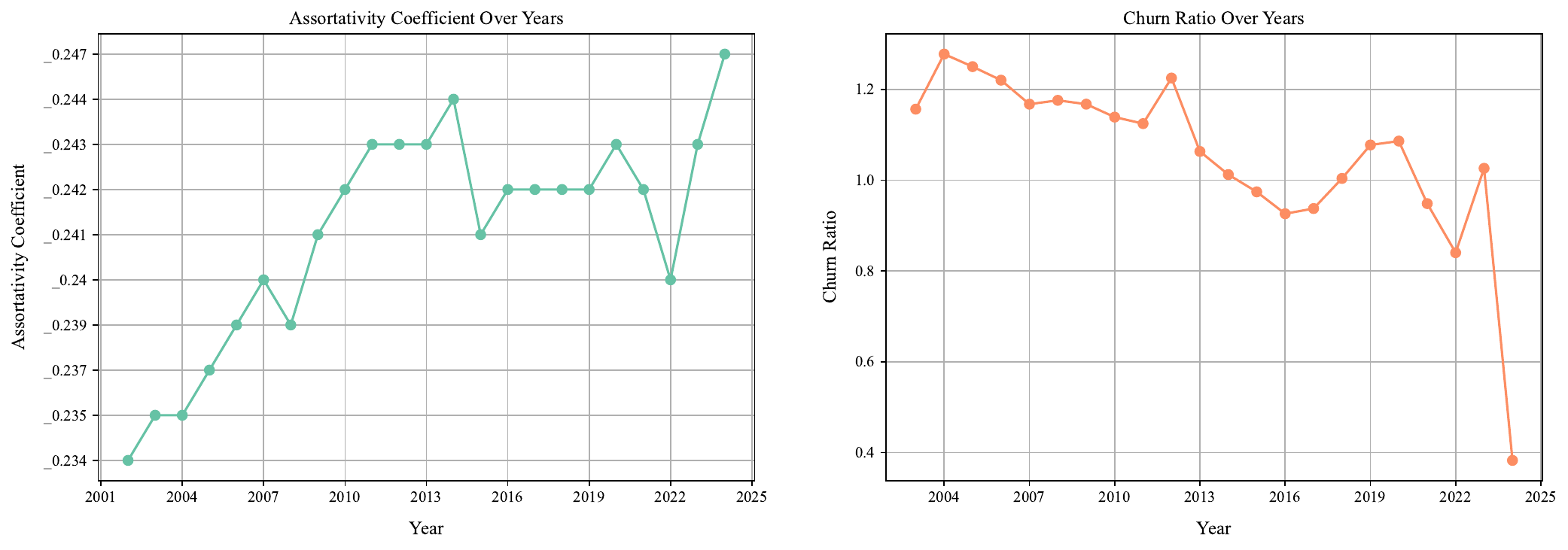}
    \caption{\textbf{FOS\textsubscript{art\&business} Temporal-Pattern Statistics:} Plots depicting (1) the assortativity coefficient over time and (2) the churn ratio (edge turnover) per year.}
    \label{fig:temporal_statistics}
\end{figure*}

We present these features through visualizations and descriptive analyses below:

\begin{enumerate}
    \item Figures~\ref{fig:node_statistics_1} and~\ref{fig:node_statistics_2} illustrate key node-level statistics for FOS\textsubscript{art\&business}. The first plot in Figure~\ref{fig:node_statistics_1} shows the number of active nodes per year, exhibiting a sharp rise from approximately 3,165 in 2001 to a peak around 3,235 by 2010, followed by minor fluctuations and a gradual decline to about 3,205 from 2021 onward. The second plot depicts the mean node degree over time, which increases steadily from around 55 in 2002 to a high of approximately 110 in 2013, before decreasing to roughly 70 by 2024. The third plot tracks the mean annual growth rate in node degrees, revealing positive growth from 2002 to 2014 (with rates up to 7), interspersed with negative rates in periods like 2015--2017, 2021--2022, and 2024, indicating intermittent contractions in node degrees. The fourth plot presents the mean clustering coefficient, starting at approximately 0.695 in 2001, dipping to around 0.645 by 2016, and then rebounding to about 0.675 by 2024. Figure~\ref{fig:node_statistics_2} complements this with distributions: the first shows node persistence spans (lifetimes in years), where the vast majority of nodes (over 3,100) remain active across the full 22-year period, with fewer nodes exhibiting shorter spans. The second distribution indicates the recency of last activity, with most nodes (3,206) last active in 2024, underscoring sustained engagement for the bulk of the graph.

    \item Figures~\ref{fig:edge_statistics_1} and~\ref{fig:edge_statistics_2} display edge-level statistics. In Figure~\ref{fig:edge_statistics_1}, the first plot tracks edge count and density over time, showing a rapid increase in edges from about 90,000 in 2002 to a peak near 180,000 around 2013, followed by a decline to approximately 120,000 by 2024; density mirrors this trend, rising from 0.009 to 0.017 before falling back. The second plot illustrates the edge repetition rate (fraction of recurring edges from prior years), which climbs from 0 in 2002 to about 0.95 by 2024, reflecting growing stability in established connections. Figure~\ref{fig:edge_statistics_2} provides distributional insights: the frequency distribution (number of years an edge appears) is heavily skewed, with around 200,000 edges appearing once and decreasing counts for higher frequencies up to 20 years. Edge lifetimes (duration from first to last appearance) reveal many long-lived edges (e.g., around 50,000 with lifetime 22) and fewer with smaller lifetimes. The mean inter-event time distribution is concentrated at low values, with most edges (near 250,000) having times near 1, tapering off beyond 5 years, indicating bursty recurrence. Finally, the last appearance year distribution shows a cumulative buildup, with the highest count (near 120,000 edges) last appearing in 2024, suggesting ongoing evolution.

    \item Figure~\ref{fig:graph_statistics} presents graph-level statistics. The first plot overlays the number of connected components (CCs) and the size of the largest CC over time. The CC count starts low at around 1 in 2002, increases sharply to 2 in 2004, and goes back to 1 thereafter. This reveals that our graph is mostly connected. Furthermore, the largest CC size grows from about 3,165 to 3,230, encompassing nearly all nodes. The second plot shows the graph diameter, which begins at 5 in 2002, fluctuates upward to 6 by 2008, and stabilizes at low values, indicating a progressively more compact network structure.

    \item Figure~\ref{fig:temporal_statistics} highlights temporal-pattern features. The assortativity coefficient (measuring temporal mixing of node degrees) evolves from -0.234 in 2002 to -0.247 by 2024, with mild fluctuations, suggesting weak disassortative tendencies where high-degree nodes connect to low-degree ones. The churn ratio (fraction of new over lost edges relative to the previous snapshot) starts below 1.2 in 2002, and declines gradually to about 0.4 by 2024, reflecting reduced volatility and novelty over the period.
\end{enumerate}

\subsection{Models Description}
\label{app:models}

Here we present more detailed descriptions of each temporal graph model considered in our study. Our goal is to clarify the architectural choices, temporal mechanisms, and memory or embedding strategies of JODIE, DyRep, TGAT, EdgeBank, GraphMixer, and DyGFormer. For each model, we explain how it processes temporal interaction data, how it updates or maintains node representations, and what trade-offs it embodies. 

\paragraph{JODIE (Joint Dynamic User–Item Embeddings)} is a continuous-time event-driven model that represents each entity via a pair of embeddings: a static embedding capturing long-term, stationary traits, and a dynamic embedding tracking time-varying behavior. At each interaction event, two mutually recursive recurrent units update the dynamic embeddings: one RNN updates the user embedding conditioned on the current item embedding, and the other updates the item embedding conditioned on the user embedding. This coupling captures the interdependence between the two sides. Beyond just reacting to events, JODIE learns a \textbf{projection operator}: given the last known embedding and the elapsed time $\delta$, it forecasts a future embedding, allowing the model to query node states at arbitrary times.  
To make training feasible at scale, JODIE employs a batching scheme called \textbf{t-Batch}, which partitions interaction edges into temporally consistent independent batches so that updates can proceed in parallel without breaking causality.

\paragraph{DyRep (Dynamic Representation learning over graphs)} frames dynamic graph evolution in terms of two intertwined temporal processes: \textbf{association} (structural or topological changes, e.g.\ new edges forming) and \textbf{communication} (interactions on existing edges). The core concept is that node embeddings mediate between these observed dynamics—representations evolve by governing and being shaped by these two event types. Technically, DyRep formulates an intensity function $\lambda(u, v, k, t)$ for a candidate event type \(k\) between nodes \(u\) and \(v\) at time \(t\). This intensity is parameterized via node embeddings and structural context, and governs both when events happen (via survival terms) and which interactions occur.  
Upon event occurrence, embeddings are updated through a combination of \textbf{self-propagation}, \textbf{structural aggregation} (attending to neighbors), and \textbf{cross-node influence} (information from the interacting partner) via an attention-like mechanism. A multiscale temporal modeling mechanism is used to allow processes on different time scales (fast communications and slower associations) to coexist. The model is trained in an unsupervised fashion by maximizing the log-likelihood of observed events minus the integral over intensities of unobserved possibilities.

\paragraph{TGAT (Temporal Graph Attention Network)} is a transformer-style temporal GNN that uses attention to aggregate information from a node’s temporal neighborhood. Concretely, for a target node at query time \(t\), TGAT considers past interacting neighbors (within a time window), encodes the time differences, and applies attention over neighbor embeddings augmented with time encodings. Thus the model can flexibly weight contributions from more recent or more distant events, and incorporate structural (graph) and temporal signals jointly. Because TGAT does not store per-node hidden states evolving by recurrence but computes embeddings on-the-fly via attention and neighbor sampling, it is naturally \textbf{inductive}: nodes unseen during training can be embedded at test time. In practice, TGAT is a strong baseline across temporal graph tasks. Its main limitations lie in the cost of neighbor sampling and attention overhead for high-degree nodes, and the risk that long-range temporal dependencies may attenuate if the attention window is truncated.

\paragraph{EdgeBank} is a memory-based heuristic baseline that forgoes complex parameterized updates, relying instead on \textbf{memorization of historical} edges to score candidate interactions. Typically, it stores observed (source, destination) pairs (or recent windows of them) and predicts that edges are likely to repeat. Two variants are common: \textbf{EdgeBank\(_\infty\)}, which retains all past edges, and \textbf{EdgeBank\(_\text{tw}\)}, which only remembers edges within a recent time window (commensurate with validation / prediction horizon). EdgeBank is extremely efficient and serves as a sanity check or lower bound: any learned dynamic model should outperform pure memorization. In many benchmarks, EdgeBank offers nontrivial performance, especially when interaction repetition is high, and is orders of magnitude faster than deep models.   

\paragraph{GraphMixer} is a simple architecture inspired by the MLP-Mixer paradigm. Rather than using graph convolutions or attention, GraphMixer maintains a memory module per node and employs MLP mixing across time or message channels to fuse information from recent interactions. It uses \textbf{fixed time encoding} functions (e.g.\ positional encodings) and mixes temporal messages with node memory via learned MLP layers. Because it eschews heavy graph operations or recurrent machinery, it is lighter, faster, and easier to scale. Despite its simplicity, GraphMixer has achieved competitive performance in dynamic graph benchmarks, suggesting that careful temporal mixing is sometimes enough to capture much of the signal. Its drawback is that it may struggle in settings where relational or graph structural nuance matters (beyond what a simple mixing can encode). 

\paragraph{DyGFormer} is a hybrid transformer-based dynamic graph model that strikes a balance: it combines transformer-style sequence modeling with structural correlation encoding and memory considerations. Given a query interaction \((u, v, t)\), DyGFormer first gathers historical \textbf{first-hop} interaction sequences of \(u\) and \(v\). It computes a \textbf{neighbor co-occurrence encoding} that captures how frequently each neighbor appears in both \(u\)’s and \(v\) ’s histories, thereby explicitly capturing correlation between their neighborhoods. Then, using a \textbf{patching technique}, each node’s sequence is divided into temporal patches, preserving local temporal continuity while reducing computational loads. All patches (for both \(u\) and \(v\)) are fed into a transformer encoder, whose output is averaged or pooled to yield time-aware node embeddings \(h_u(t)\), \(h_v(t)\). This architecture enables learning from \textbf{long histories} without prohibitive complexity, and captures \textbf{correlations across nodes} in interactions. In experiments across many datasets, DyGFormer achieves SOTA performance on dynamic link prediction and node classification tasks. Its trade-off is increased computational cost and more hyperparameter tuning compared to lighter models.  

\subsection{Metrics Description}
\label{app:metrics}

We evaluate predictive performance using two complementary metrics that are robust to severe class imbalance and sensitive to ranking quality. Average Precision (AP) is computed per test batch from the ranked predicted probabilities and subsequently averaged across all batches and test years. Formally, if $P_n$ and $R_n$ denote the precision and recall at the $n$-th distinct score threshold, AP is computed as
\begin{equation}
\mathrm{AP} \;=\; \sum_{n=1}^{N} (R_n - R_{n-1}) \cdot P_n, \quad P_0 = R_0 = 0
\end{equation}
where $N$ is the number of thresholds. 

The Area Under the Receiver Operating Characteristic curve (AUC--ROC) complements AP by measuring the probability that a randomly chosen positive example is scored higher than a randomly chosen negative example. AUC--ROC is computed per batch and averaged across batches and test years to produce a single aggregate score. All metric computations use the continuous predicted probabilities and follow standard numerical implementations (from \texttt{scikit-learn}) to ensure reproducibility.

\subsection{Negative Sampling Strategies}
\label{app:negative_sampling}

We adopt three negative-sampling regimes that produce progressively harder evaluation sets and that isolate distinct generalization challenges. Let $V$ denote the node set, $E_{\mathrm{train}}$ the training-edge set, $E_{\mathrm{test}}$ the test-edge set, and $E_t$ the set of edges observed at time $t$. For a positive edge $(u,v,t)$, the \emph{random} sampling regime draws negative destination nodes $v'\in V\setminus\{v\}$ uniformly at random while preserving the source node $u$, timestamp $t$, and any auxiliary features associated with the positive edge. This strategy evaluates transductive discrimination among known nodes. The \emph{historical} regime samples negatives from previously observed edges that are absent at the current step,
\begin{equation}
\mathcal{N}_{\mathrm{hist}}(t) = \{(u,v',t)\;|\;(u,v')\in E_{\mathrm{train}}\cap E_t'\}
\end{equation}

and therefore probes a model's ability to distinguish true future re-occurrences from spurious recollections of past interactions. The \emph{inductive} regime draws negatives from edges that only appear during the test period,
\begin{equation}
\mathcal{N}_{\mathrm{ind}}(t) = \{(u,v',t)\;|\;(u,v')\in \left(E_{\mathrm{test}}\setminus E_{\mathrm{train}}\right) \cap E_t'\}
\end{equation}
thereby measuring generalization to novel link patterns not available during training.

In this work, we sample one negative per positive by default. When the historical or inductive pool is insufficient to match the number of positives in a batch, the remaining negatives are supplemented by draws from the random pool. This fallback mechanism preserves the intended difficulty ordering (random $\rightarrow$ historical $\rightarrow$ inductive) while guaranteeing a full set of negatives for metric computation. The sampling procedure is implemented deterministically given a fixed random seed to ensure run-to-run reproducibility.

\subsection{Neighbor Sampling Strategies}
\label{app:neighbor_sampling}

Temporal neighborhood construction is central to capturing evolving local structure. Let node $u$ have historical neighbor interactions $\{(n_i,t_i) \,|\, (u, n_i) \in E_{t_i}\}_{i=1}^M$ with timestamps $t_i<t$. We consider three sampling schemes. Uniform sampling selects up to $S$ neighbors uniformly at random from the available set, thereby treating all past interactions equally. Recent sampling deterministically selects the $S$ most recent neighbors, which concentrates information on the most temporally proximate interactions. Time-interval-aware sampling draws neighbors without replacement according to an exponential recency bias:
\begin{equation}
\mathbb{P}(n_i) \;=\; \frac{\exp\big(-\alpha\,(t - t_i)\big)}{\sum_{j=1}^{M}\exp\big(-\alpha\,(t - t_j)\big)},
\end{equation}
where $\alpha>0$ is a time-scaling factor that controls the degree of recency preference. Sampling proceeds until $S$ neighbors are selected or all available interactions are exhausted; when fewer than $S$ neighbors exist we pad the aggregation input with a fixed null embedding so that downstream layers receive a constant-size input. In our experiments we report results for $S=20$ and use $\alpha=1\times 10^{-6}$ for the time-interval-aware variant.

Neighbor sampling is performed on-the-fly when forming minibatches so that only interactions strictly preceding the prediction time $t$ are visible to the model. All stochastic samplers are seeded to permit deterministic re-execution of experiments. Tables~\ref{tab:main_results_uniform} and~\ref{tab:main_results_time} report the results for the two neighbor-sampling strategies we evaluated: Uniform and time-interval-aware, respectively.

\subsection{Implementation Details}
\label{app:implementation}

All experiments are conducted on an Art+Business subset of the FOS benchmark that includes descendant subfields and comprises $N=3{,}238$ nodes and 3,472,315 edges. Temporal splits are chronological: training uses edges observed between 2002 and 2017, validation uses edges from 2018 to 2021, and testing covers 2022 to 2024. This partitioning enforces a strict temporal separation between training and evaluation and tests the models' ability to predict future interdisciplinary links.

Each node is represented by a semantic feature vector \(e_{v}\in\mathbb{R}^{768}\) (see Section~\ref{proposed}). We applied Principal Component Analysis (PCA) to reduce these vectors from 768 to 100 dimensions for all nodes.

Model architectures share a common scoring head: given node embeddings $z_u$ and $z_v$ produced by a temporal encoder up to time $t$, link probability is computed as $\hat{p}_{uv}=\mathrm{MLP}([z_u\Vert z_v])$, where $[\cdot\Vert\cdot]$ denotes concatenation and the MLP terminates in a sigmoid activation. The MLP topology and other model hyperparameters are selected according to validation AP; the reported runs use a compact two-layer MLP with ReLU activations.

A concise summary of the principal hyperparameters used in the reported experiments follows. Models employ two graph neural network layers with two attention heads per layer. Neighborhood sampling uses $S=20$ neighbors; random-walk–based aggregation uses walk length one with eight walk heads. The time feature is embedded in a 100-dimensional space and positional embeddings are 172-dimensional. EdgeBank is configured with \texttt{unlimited\_memory} and a \texttt{fixed\_proportion} time window mode. The channel embedding dimension is 50, patch size is 1, and the maximum input sequence length is 32. Training uses the Adam optimizer with a learning rate of $1\times 10^{-4}$, a dropout rate of 0.1, no weight decay, and a maximum of 30 epochs; early stopping with patience 20 is applied to prevent overfitting. Batches are formed chronologically with batch size 300. 

To facilitate reproducibility, we fix random seeds for Python, NumPy, and PyTorch. All the fitted transformers are saved alongside model checkpoints. Model selection is based on validation AP; final test results are computed using the selected checkpoint and the identical preprocessing pipeline. Metric computations use continuous probabilities and standard implementations (from \texttt{scikit-learn}) to ensure numerical consistency.

All experiments were conducted on a system running Ubuntu 22.04.5 LTS, equipped with a 13th Gen Intel Core i7-13700K CPU, 64GB of RAM, and an Nvidia GeForce RTX 4090 GPU. This hardware configuration ensured efficient processing of the large-scale FOS\textsubscript{art \& business} dataset and supported the computational demands of training and evaluating temporal Graph Neural Network models.

\paragraph{Code and Data Availability.}
The repository accompanying this work provides scripts to (i) reproduce the dataset splits used for the Art+Business subset, (ii) fit and persist preprocessing transforms, (iii) train candidate models, and (iv) evaluate models under the three negative-sampling and three neighbor-sampling regimes. The repository also records all hyperparameter settings, random seeds, and the exact dataset version used for each reported run. 




\subsection{List of Predictions}
\label{app:predictions}

We have demonstrated how the FOS model's predictions can be useful in real-world applications. Here, we extend the results presented in Table~\ref{tab:real-world} to provide a broader overview of high-confidence predicted cross-field links (see Table~\ref{tab:real-world(continue)}).

\subsection{TEA and TET Plots}
\label{app:tea_tet}

Accompanying the framework, three metrics—novelty, re-occurrence, and surprise—were reported in Table~\ref{tab:novelty_recurrence_surprise} to quantify the temporal edge dynamics of our benchmark. Here, two plots are presented to further elucidate these metrics:

\begin{itemize}
    \item \textbf{Temporal Edge Appearance (TEA) Plot:}
    This plot illustrates the distribution of new and repeated edges over time, providing insight into the temporal dynamics of novelty.
    \item \textbf{Temporal Edge Traffic (TET) Plot:}
    This plot depicts edge appearances across different timestamps, offering a visual representation of the re-occurrence and surprise of edges over time. The edges are sorted based on their first appearance (edges that appear at the same timestamp are then sorted based on their last appearance). Edges seen in training are colored in green and edges seen only in test (inductive edges) are colored in red.
\end{itemize}

\begin{figure*}[!t]
    \centering
    \includegraphics[width=0.95\linewidth]{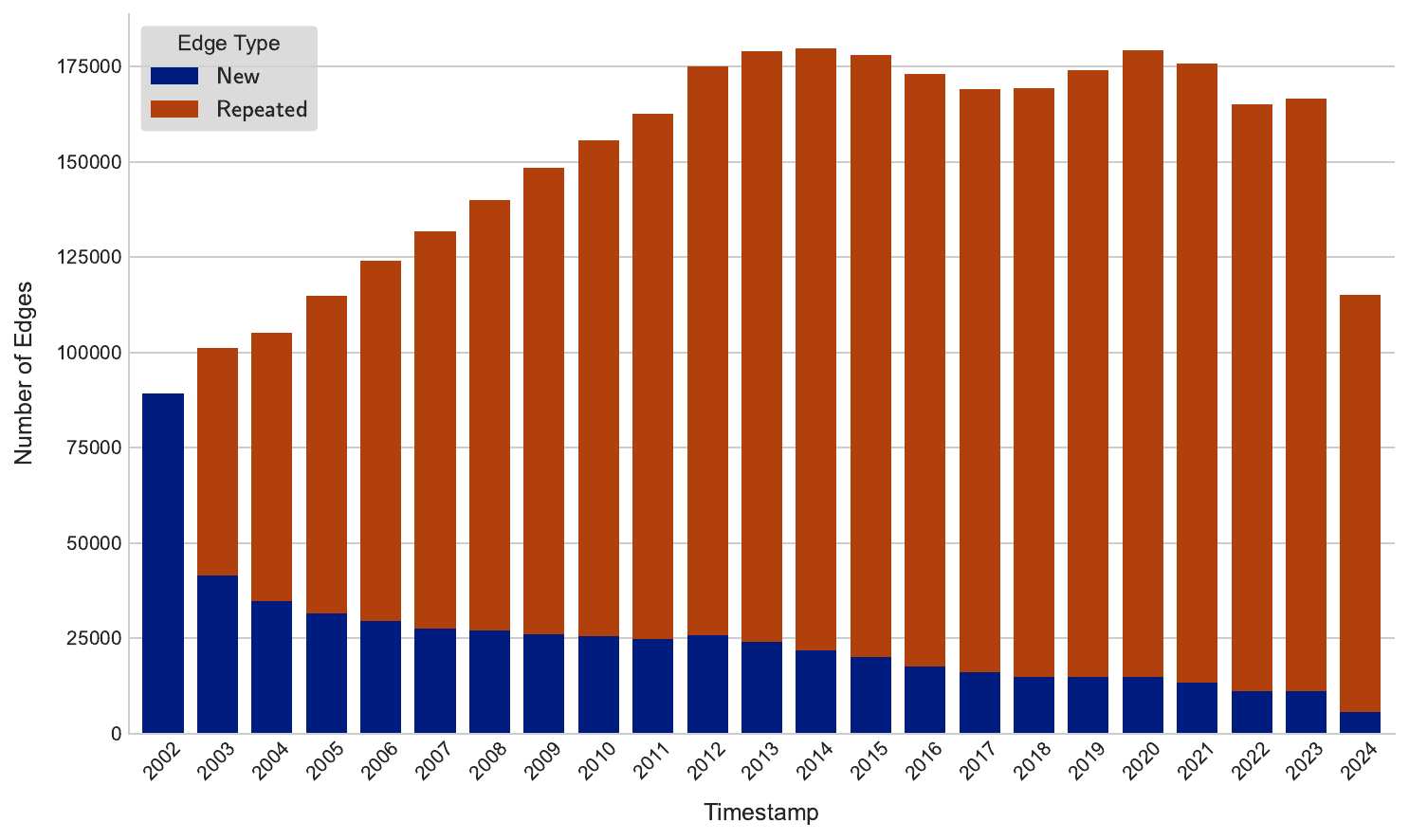}
    \caption{\textbf{Temporal Edge Appearance (TEA) Plot}. Timeline of edge appearances sorted by first occurrence; colors distinguish edges seen during training (repeated) from edges that first appear in the test period (new), visualizing the temporal distribution of novelty and recurrence.}
    \label{fig:fig1}
    
    \vspace{0.5cm} 
    
    \includegraphics[width=0.95\linewidth]{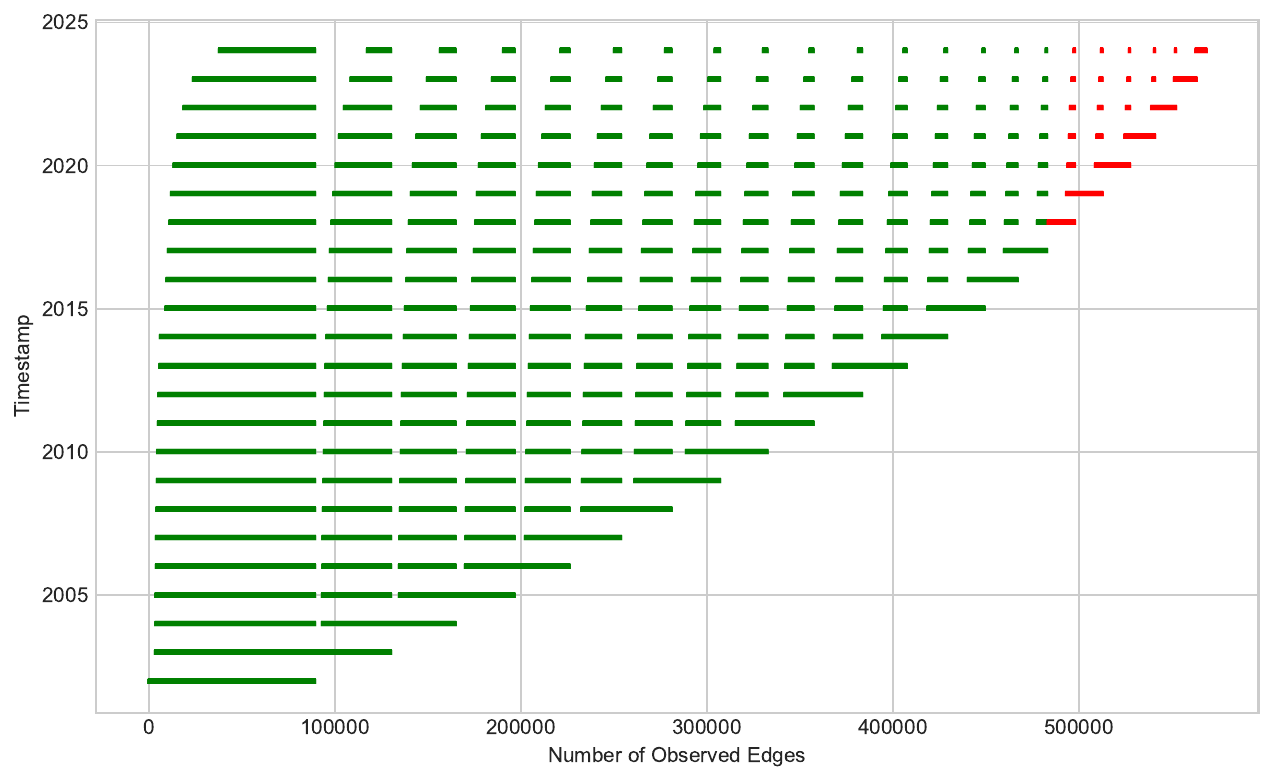}
    \caption{\textbf{Temporal Edge Traffic (TET) Plot}. Edge-appearance heatmap across timestamps that highlights traffic density and separates inductive (test-only) edges (red) from training-seen edges (green) to illustrate surprise and re-occurrence patterns.}
    \label{fig:fig2}
\end{figure*}

Figures \ref{fig:fig1} and \ref{fig:fig2} correspond to these plots, respectively.

\begin{table*}[h]
\centering
\renewcommand{\arraystretch}{1.2} 
\setlength{\tabcolsep}{12pt} 
\begin{tabular}{l|cc|cc|cc}
\toprule
\multirow{2}{*}{\textbf{Model}} & \multicolumn{2}{c|}{\textbf{Random}} & \multicolumn{2}{c|}{\textbf{Inductive}} & \multicolumn{2}{c}{\textbf{Historical }} \\
& \textbf{AP} & \textbf{AUC-ROC} & \textbf{AP} & \textbf{AUC-ROC} & \textbf{AP} & \textbf{AUC-ROC} \\
\midrule
JODIE        & 0.7572 & 0.7594 & 0.6485 & 0.6497 & 0.5671 & 0.5655 \\
EdgeBank     & 0.7697 & 0.8412 & 0.6453 & 0.6442 & 0.4852 & 0.4689 \\
DyRep        & 0.8442 & 0.8623 & 0.7214 & 0.7563 & 0.6019 & 0.6314 \\
GraphMixer   & 0.8001 & 0.8277 & 0.7689 & 0.7849 & 0.6468 & 0.6617 \\
TGAT         & 0.8807 & 0.8856 & 0.7930 & 0.8083 & 0.6378 & 0.6559 \\
DyGFormer    & 0.8898 & 0.8972 & 0.7094 & 0.7358 & 0.6015 & 0.6226 \\
\bottomrule
\end{tabular}
\vspace{2mm}
\caption{\textbf{Main results on FOS\textsubscript{art\&business} using uniform neighbor sampling.} 
We report Average Precision (AP) and AUC-ROC under three evaluation regimes: \emph{Random}, \emph{Inductive}, and \emph{Historical}. 
Best scores in each column are shown in bold.}
\label{tab:main_results_uniform}
\end{table*}

\begin{table*}[h]
\centering
\renewcommand{\arraystretch}{1.2} 
\setlength{\tabcolsep}{12pt} 
\begin{tabular}{l|cc|cc|cc}
\toprule
\multirow{2}{*}{\textbf{Model}} & \multicolumn{2}{c|}{\textbf{Random}} & \multicolumn{2}{c|}{\textbf{Inductive}} & \multicolumn{2}{c}{\textbf{Historical}} \\
& \textbf{AP} & \textbf{AUC-ROC} & \textbf{AP} & \textbf{AUC-ROC} & \textbf{AP} & \textbf{AUC-ROC} \\
\midrule
JODIE        & 0.7572 & 0.7594 & 0.6485 & 0.6497 & 0.5671 & 0.5655 \\
EdgeBank     & 0.7697 & 0.8412 & 0.6453 & 0.6442 & 0.4852 & 0.4689 \\
DyRep        & 0.8449 & 0.8627 & 0.7229 & 0.7581 & 0.6022 & 0.6314 \\
GraphMixer   & 0.8000 & 0.8276 & 0.7689 & 0.7849 & 0.6470 & 0.6617 \\
TGAT         & 0.8805 & 0.8856 & 0.7931 & 0.8086 & 0.6377 & 0.6560 \\
DyGFormer    & 0.8898 & 0.8972 & 0.7094 & 0.7358 & 0.6015 & 0.6226 \\
\bottomrule
\end{tabular}
\vspace{2mm}
\caption{\textbf{Main results on FOS\textsubscript{art\&business} using time-interval-aware neighbor sampling.}
We report Average Precision (AP) and AUC-ROC under the \emph{Random}, \emph{Inductive}, and \emph{Historical} evaluation regimes.
Best scores in each column are shown in bold.}
\label{tab:main_results_time}
\end{table*}


\begin{center}
\onecolumn 
\setlength{\LTpre}{0pt}
\setlength{\LTpost}{0pt}

\begin{longtable}{>{\raggedright\arraybackslash}p{8.75cm} >{\centering\arraybackslash}p{1.75cm} >{\centering\arraybackslash}p{1.5cm} >{\raggedright\arraybackslash}p{5cm}}
\caption{\textbf{Extended Research Data (Selected Predicted Links).}}\\
\toprule
\arrayrulecolor{black}
Title & Publication Date & Prediction Score & Concepts \\
\midrule
\endfirsthead


\arrayrulecolor{black}
\bottomrule
\endfoot

\endlastfoot

\arrayrulecolor{gray!30}

From Scroll to Purchase: Influencing Tourist Travel Purchase on Travel E-Commerce \href{https://openalex.org/W4409341759}{W4409341759} & 2025-04-10 & \num{0.9545126} & Travel writing - Commerce \\
\arrayrulecolor{gray!30}\midrule
THE WORLD THEATRE OF ANTON CHEKHOV BOOK REVIEW: KISSEL V.S. CHEKHOV’S COSMOS. THEATRE, SPACE AND TIME. SAINT PETERSBURG: ALETEYA, 2024. 410 P. \href{https://openalex.org/W4410646743}{W4410646743} & 2025-01-01 & \num{0.9514137} & Cosmos (plant) - Russian federation \\
\arrayrulecolor{gray!30}\midrule
Research on the Restructuring of the Visual Communication Design Curriculum System Driven by Management Principles \href{https://openalex.org/W4412418263}{W4412418263} & 2025-06-13 & \num{0.94399863} & Communication design - Restructuring \\
\arrayrulecolor{gray!30}\midrule
OS-042 Benchmarking spleen stiffness in assessing response to betablocker therapy for secondary prophylaxis of acute variceal bleed-BE-RESPONSE study (NCT05166317) \href{https://openalex.org/W4410175234}{W4410175234} & 2025-05-01 & \num{0.9438082} & Bleed - Benchmarking \\
\arrayrulecolor{gray!30}\midrule
\textit{BeautyClicker}: Modeling Viewers’ Click Intent of the Cover Image to Support Drafts of Beauty Product Promotion Posts \href{https://openalex.org/W4412641408}{W4412641408} & 2025-07-24 & \num{0.9368667} & Click-through rate - Beauty \\
\arrayrulecolor{gray!30}\midrule
Specific media literacy tips improve AI-generated visual misinformation discernment \href{https://openalex.org/W4412004671}{W4412004671} & 2025-07-03 & \num{0.93596804} & Legibility - Media literacy \\
\arrayrulecolor{gray!30}\midrule
Comparative Analysis of Performance and Cost of Chemical Stabilizers for Iowa Granular Roads: Field and Laboratory Evaluation \href{https://openalex.org/W4410239797}{W4410239797} & 2025-05-09 & \num{0.93421376} & Gradation - Cost effectiveness \\
\arrayrulecolor{gray!30}\midrule
Internationalization at Home: Activities on Foodways in Spanish Cultural Studies Classes \href{https://openalex.org/W4411100412}{W4411100412} & 2025-06-01 & \num{0.93121386} & Foodways - Internationalization \\
\arrayrulecolor{gray!30}\midrule
A Social Hierarchy Perspective on the Detrimental Effects of Leader–Member Exchange Differentiation on Team Functioning: Leader‐Conferred Status Versus Member‐Conferred Status \href{https://openalex.org/W4413807266}{W4413807266} & 2025-08-29 & \num{0.93044436} & Social hierarchy - Member states \\
\arrayrulecolor{gray!30}\midrule
The Effect of Education Management and Learning and Teaching Arts on Organizational Wellbeing: The Moderating Role of Organizational Support in Tertiary Level Education \href{https://openalex.org/W4409277575}{W4409277575} & 2025-04-06 & \num{0.92712086} & Organizational behavior and human resources - The arts \\
\arrayrulecolor{gray!30}\midrule
Exploring the Intersections of Bildung and Social Entrepreneurship: Implications for Entrepreneurship Education in Korea \href{https://openalex.org/W4413203622}{W4413203622} & 2025-07-31 & \num{0.9229939} & Bildung - Social entrepreneurship \\
\arrayrulecolor{gray!30}\midrule
Predicting the future of the corporate market: a proposed dual fraud-bankruptcy score based on evidence from Romanian companies \href{https://openalex.org/W4409177342}{W4409177342} & 2025-04-01 & \num{0.91542155} & Bankruptcy prediction - Dual (grammatical number) \\
\arrayrulecolor{gray!30}\midrule
Examining the Influence on Organ Donation by Actor Portrayal of Altruism on the Television Show \textit{Grey's Anatomy} \href{https://openalex.org/W4410619512}{W4410619512} & 2025-05-23 & \num{0.91478056} & Conflict of interest - Realism \\
\arrayrulecolor{gray!30}\midrule
Enhancing Learning On The Brand Building Of Culture And Creative Industries Companies Integrating Brand Identity System And Creative Supply Chain Theories \href{https://openalex.org/W4410488798}{W4410488798} & 2025-05-19 & \num{0.9123293} & Creative industries - Brand identity \\
\bottomrule
\label{tab:real-world(continue)}
\end{longtable}
\end{center}

\end{document}